\documentclass{article}

\usepackage{microtype}
\usepackage{graphicx}
\usepackage{subfigure}
\usepackage{booktabs} 

\usepackage{hyperref}

\usepackage[accepted]{icml2024}

\usepackage{amsmath}
\usepackage{amssymb}
\usepackage{mathtools}
\usepackage{amsthm}

\usepackage{color}
\usepackage{multirow}
\usepackage{wrapfig}
\usepackage{tabularx}
\usepackage{hyperref}
\usepackage{url}

\newtheorem{observation}{Observation}

\DeclareMathOperator*{\argmin}{argmin}

\usepackage[capitalize,noabbrev]{cleveref}

\theoremstyle{plain}
\newtheorem{theorem}{Theorem}[section]
\newtheorem{proposition}[theorem]{Proposition}

\theoremstyle{definition}
\newtheorem{definition}[theorem]{Definition}

\theoremstyle{remark}

\usepackage[textsize=tiny]{todonotes}

\icmltitlerunning{Transitional Uncertainty with Layered Intermediate Predictions}

\begin{document}

\onecolumn 

\begin{description}

\item[\textbf{Citation}] R. Benkert, M. Prabhushankar, and G. AlRegib, “Transitional Uncertainty with Layered Intermediate Predictions,” in International Conference on Machine Learning (ICML), Vienna, Austria, 2024 \\


\item[\textbf{Review}]
{
Date of submission: 29 February 2024\\
Date of revision: 28 March 2024\\
Date of acceptance: 1 May 2024
} \\


\item[\textbf{Bib}] {@inproceedings\{benkert2024\_TULIP,\\ 
author=\{R. Benkert, M. Prabhushankar, and G. AlRegib\},\\ 
journal=\{International Conference on Machine Learning (ICML)\},\\ 
title=\{Transitional Uncertainty with Layered Intermediate Predictions\}, \\ 
year=\{2024\}\}\\ 
} \\


\item[\textbf{Copyright}]{The authors grant ICML a non-exclusive, perpetual, royalty-free, fully-paid, fully-assignable license to copy, distribute and publicly display all or parts of the paper. Personal use of this material is permitted. Permission from the authors must be obtained for all other uses, in any current or future media, including reprinting/republishing this material for advertising or promotional purposes, for resale or redistribution to servers or lists.}
\\
\item[\textbf{Contact}]{\href{mailto:rbenkert3@gatech.edu}{rbenkert3@gatech.edu}  OR \href{mailto:alregib@gatech.edu}{alregib@gatech.edu}\\ \url{http://ghassanalregib.info/} \\ }

\end{description}

\thispagestyle{empty}
\newpage
\clearpage
\setcounter{page}{1}

\twocolumn
\twocolumn[
\icmltitle{Transitional Uncertainty with Layered Intermediate Predictions}



\icmlsetsymbol{equal}{*}

\begin{icmlauthorlist}
\icmlauthor{Ryan Benkert}{yyy}
\icmlauthor{Mohit Prabhushankar}{yyy}
\icmlauthor{Ghassan AlRegib}{yyy}
\end{icmlauthorlist}

\icmlaffiliation{yyy}{School of Electrical and Computer Engineering, Georgia Institute of Technology, Atlanta, USA}


\icmlcorrespondingauthor{Ryan Benkert}{rbenkert3@gatech.edu}

\icmlkeywords{Single-Pass Uncertainty Estimation, Out-of-Distribution Detection}

\vskip 0.3in
]



\printAffiliationsAndNotice{}  

\begin{abstract}
In this paper, we discuss feature engineering for single-pass uncertainty estimation. For accurate uncertainty estimates, neural networks must extract differences in the feature space that quantify uncertainty. This could be achieved by current single-pass approaches that maintain feature distances between data points as they traverse the network. While initial results are promising, maintaining feature distances within the network representations frequently inhibits information compression and opposes the learning objective. We study this effect theoretically and empirically to arrive at a simple conclusion: preserving feature distances in the output is beneficial when the preserved features contribute to learning the label distribution and act in opposition otherwise. We then propose \emph{Transitional Uncertainty with Layered Intermediate Predictions} (\texttt{TULIP}) as a simple approach to address the shortcomings of current single-pass estimators. Specifically, we implement feature preservation by extracting features from intermediate representations before information is collapsed by subsequent layers. We refer to the underlying preservation mechanism as \emph{transitional feature preservation}. We show that \texttt{TULIP} matches or outperforms current single-pass methods on standard benchmarks and in practical settings where these methods are less reliable (imbalances, complex architectures, medical modalities). 
\end{abstract}

\vspace{-3mm}
\section{Introduction}

Effective single-pass uncertainty estimation in deep learning is governed by two design principals. The first is defining an output score that reflects uncertainty. For instance, we can measure uncertainty through distance from the training data \citep{liu2020simple}, or the softmax confidence of the output \citep{mukhoti2023deep}. While score design plays a crucial role in measuring uncertainty, the choice is dictated by application and uncertainty characteristics \citep{kendall2017uncertainties}. The second principle concerns information availability, namely whether the network can preserve features that reflect uncertainty information and does not ``collapse" uncertain data points to certain representations \citep{van2020duq}. We refer to the latter principle as \emph{feature preservation}.

\begin{figure*} [ht]
    \begin{center}
        \includegraphics[scale=0.4]{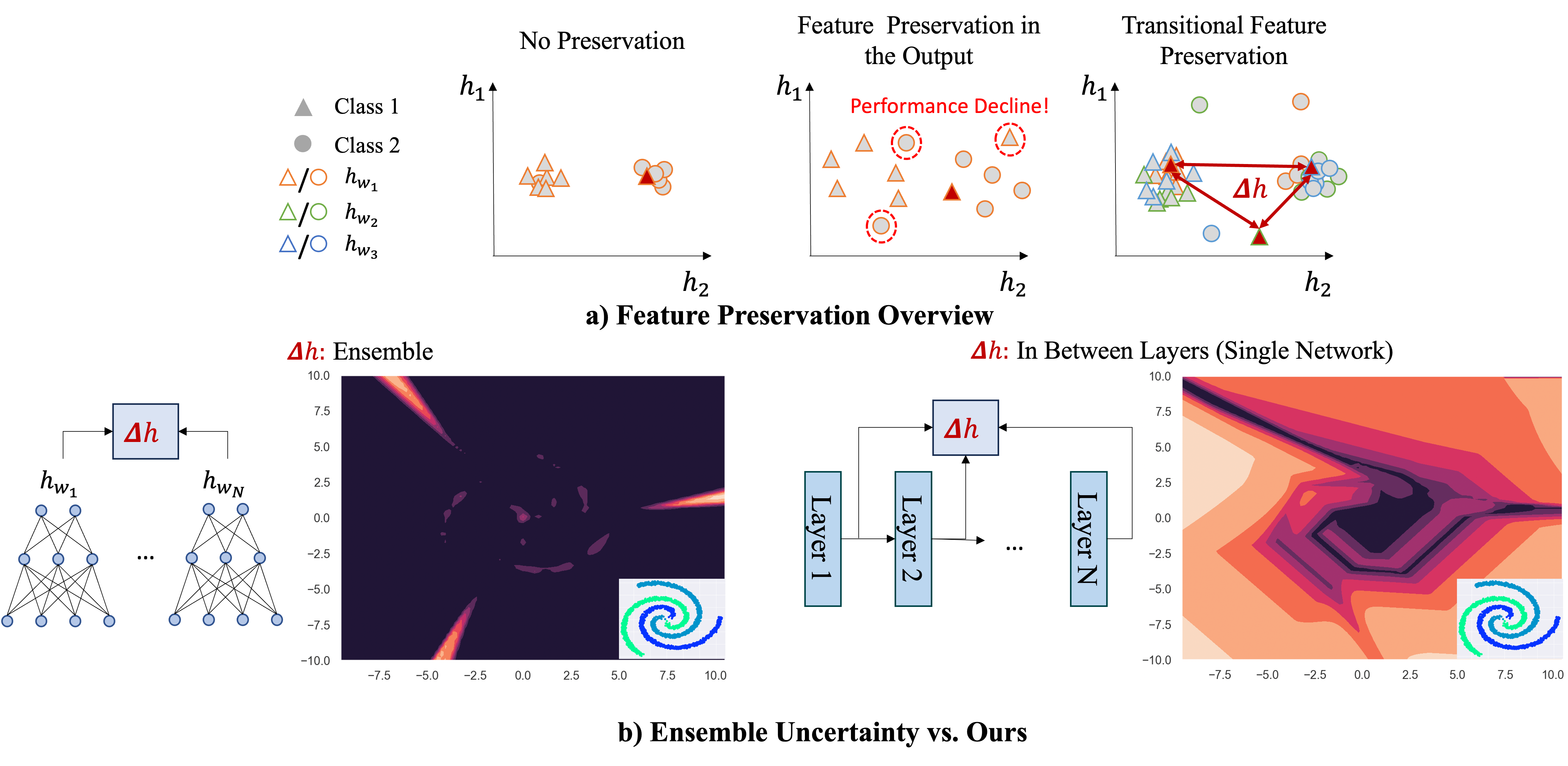}
    \end{center}
    \vspace{-3mm}
    \caption{\textbf{a)} overview of different feature preservation paradigms. We show 2D representations of neural networks where $h_1$ and $h_2$ denote the output dimensions in the feature space. Left: conventional neural network. Samples are collapsed to two tight clusters with little uncertainty information. Center: feature preservation in the output. Feature differences are maintained resulting in higher uncertainty related content but also in a cluster overlap. Right: transitional feature preservation. uncertainty is measured from differences between several representations of the same sample (denoted as $h_{w_{1,2,3}})$. \textbf{b)} uncertainty comparison of our intuition to ensembles. Transitions in between network layers provide an accurate signal for uncertainty estimation in comparison to ensembles.}
    \label{fig:intro-overview}
\end{figure*}

Despite their critical importance, preserving features is not trivial in neural networks. In particular, information compression is a desirable property of neural networks and a central component of the learning problem \citep{tishby2000information}. In spite of this discrepancy, current single-pass uncertainty methods preserve features by maintaining distances between data points in the output and risk inhibiting compression of application irrelevant information \citep{liu2020simple, van2020duq, van2021due, mukhoti2023deep, kwon2020backpropagated, prabhushankarintrospective}. As a result, several recent studies have shown practical limitations of current single-pass methods such as their susceptibility to distributional shift \citep{postels22apracticality}. We provide a simple illustration of this effect in Figure~\ref{fig:intro-overview}a. The left plot shows the 2D neural network output features of two clusters when trained without explicit feature preservation. The network collapses the class clusters to single points creating a challenging setting for uncertainty estimation. The center plot shows the same 2D classification problem, but depicts a network trained with feature preservation constraints on the output. These disadvantages motivate the search for alternative feature preservation approaches. 

To this end, a more conservative strategy preserves features in the collection of multiple representations of the same sample \citep{malinin2018predictive}. From a preservation perspective, the approach is preferred as feature distances are encoded in differences (e.g. spread) of the individual sample representations and does not require constraining the compression property. We term this approach as \emph{transitional feature preservation} and illustrate a toy example in the right plot of Figure~\ref{fig:intro-overview}a. We show three different representations of the same two clusters collected from different sources $h_{w_{1,2,3}}$. While the sources collapse points individually, the inter-source difference for a given sample reflects uncertainty (signified by $\Delta h$). A classic candidate that could implement this property are deep ensembles \citep{lakshminarayanan2017simple}, where each source representation is collected from a separate network. However, ensembles face two central challenges for uncertainty estimation: 1) the evaluation requires several forward passes limiting their practicality; and 2) there is no guarantee that ensemble transitions preserve features because all ensemble members can collapse features to the same point collectively. We show an example of this phenomenon with a synthetic 2D spiral dataset (Figure~\ref{fig:intro-overview}b left plot). Uncertainties collected from ensemble transitions fail to capture uncertainty far away from the training data even though multiple networks are used to preserve features. Our method addresses both of these limitations. We provide experimental details and further discussion of our synthetic example in Appendix~\ref{app:exp-setup-art-dataset} and \ref{app:disagreement-weight-tuning} respectively.

In this paper, we address shortcomings of current single-pass uncertainty estimators and ensembles by leveraging intermediate representations. Figure~\ref{fig:intro-overview}b (right plot) shows the intuition behind our approach. Features are extracted from intermediate layers before information is collapsed allowing the uncertainty calculation within a single forward pass. From the uncertainty plot, we can see the advantage of our approach in comparison to ensembles. Uncertainty increases systematically further away from the training data showing while ensembles are uncertain only around decision boundaries. In addition to empirical evidence, we provide theoretical guarantees for intermediate representations in Section~\ref{sec:methodology}. In particular, we show that the linear combination of representations is feature preserving when the first layer is collapse resistant. Finally, we combine intermediate representations with a single-pass uncertainty estimation layer, approximate Gaussian Processes, and arrive at a new single-pass model: \emph{Transitional Uncertainty with Layered Intermediate Predictions} or \texttt{TULIP} in short. Our estimator requires less labeled training data and outperforms current single-pass estimators both on standard benchmarks and other data modalities (CT scans) with little additional space overhead. Further, we show that \texttt{TULIP} is preferable under complex architectures, class imbalance, and several other challenging settings for current single-pass estimators.




\vspace{-3mm}
\section{Background}

\subsection{Neural Networks in the Information Plane}
\label{sec:dnn-information-plane}

Consider the input space $\mathcal{X}$ with a corresponding probabilistic random variable $X$. Further, let $\mathcal{Y}$ denote a lower dimensional target space characterized by variable $Y$. The learning problem for neural networks is equivalent to finding the minimally sufficient statistical mapping $h^*(X)$ with respect to the mutual information $I(X; Y)$ \cite{shwartz2017opening}.

\vspace{-3mm}
\begin{equation}
\label{eq:minimal-sufficient-statistic}
\begin{split}
   h^*(X) = \argmin_{h_w: I(h_w(X); Y) = I(X; Y)} I(h_w(X); X)
\end{split}
\end{equation}

\vspace{-2mm}
Equation~\ref{eq:minimal-sufficient-statistic} is intuitive. During training, we optimize the network $h_w$ to fit the lower dimensional distribution $Y$ - i.e. maximize the mutual information $I(h_w(X); Y)$ between the representation distribution $h_w(X)$ and the target distribution $Y$. At the same time, the neural network must compress information irrelevant to the lower dimensional target variable $Y$. In Equation~\ref{eq:minimal-sufficient-statistic}, we minimize the mutual information $I(h_w(X); X)$ between the representation $h_w(X)$ and the input distribution $X$. In practice, we derive $h_w$ from training data $D = \{y_i, x_i\}_{i=1}^{N}$ often collected from a subset of the full input space $\mathcal{X}_{ID} \subset \mathcal{X}$. As a result, the network optimizes with respect to the in-distribution variable $X_{ID}$ and produces arbitrarily bad results when exposed to out-of-distribution (OOD) data $\mathcal{X}_{OOD} \subset \mathcal{X}: \mathcal{X}_{OOD} \cap \mathcal{X}_{ID} = \emptyset$. For this reason, accurate uncertainty estimation is contingent on \emph{modeling information related to the full input distribution} without over-fitting to $\mathcal{X}_{ID}$ \citep{liu2020simple}.

\subsection{Distance-Based Feature Preservation in the Output}

An intuitive approach to uncertainty estimation involves modeling distributional information in the output of the neural network feature extractor \citep{van2020duq, van2021due, mukhoti2023deep, liu2020simple}. The approach is practical as we can compute the network output in a single forward pass and measure uncertainty from the logits directly. To ensure accurate uncertainty estimates, current single-pass methods model input information by maintaining the distances between data points as they traverse the network. By preserving meaningful distances, we can estimate uncertainty by measuring the distance to the training domain $\mathcal{X}_{ID}$ \citep{liu2020simple}. More formally, given an input space $\mathcal{X}$ equipped with a meaningful distance $d_X$, we learn a neural network $h_w: \mathcal{X} \rightarrow \mathcal{H}$ that allows a distance $d_H$ within the feature manifold that reflects the true distance $d_X$ \citep{liu2020simple}:

\vspace{-4mm}
\begin{equation}
\label{eq:representational-feature-preservation}
\begin{split}
   d_H(h_w(\mathbf{x}_1), h_w(\mathbf{x}_2)) = d_X(\mathbf{x}_1, \mathbf{x}_2)
\end{split}
\end{equation}
Unfortunately, neural networks do not naturally implement distance preservation and ``collapse" data points to the same output. For this purpose, current single-pass methods enforce distance preservation artificially through constraints on the network representation. Popular examples of constraints include the two-sided gradient penalty \citep{gulrajani2017improved} and spectral normalization in combination with residual connections \citep{miyato2018spectral}.

\vspace{-3mm}
\section{Theoretical Analysis}
\label{sec:theory-overall}

\subsection{Pitfalls of Feature Preservation in the Output}

While distance preservation in the output is a desirable property for uncertainty estimation, enforcing Equation~\ref{eq:representational-feature-preservation} frequently results in performance degradation in neural networks: directly preserving distances in the network output can inhibit compression of information; a learning objective according to Equation~\ref{eq:minimal-sufficient-statistic}. In this section, we provide theoretical justification that enforcing distance preservation on the network can act in direct opposition to the learning problem. We further arrive at an intuitive conclusion: distance preservation in the output is beneficial only when the preserved distances contain information related to the label distribution $Y$ - i.e. when they are relevant to the application.

We start our discussion by connecting the learning problem in Equation~\ref{eq:minimal-sufficient-statistic} to distances in the feature space. In particular, the minimally sufficient statistic shares the following dependency to feature distances for networks preserving distances in the output:
\vspace{-3mm}

\begin{equation}
\label{eq:minimally-sufficient-statistic-distances}
\begin{split}
   h^*(X) &= \argmin_{h_w: \{ I(f_H^k(h_w(M^k)); Y^{k}) = I(M^k; Y^{k}),~k \in [1, N_p]\} } \\
   &\sum_{k} I(f_H^k(h_w(M^k)); f_X^k(M^k)).
\end{split}
\end{equation}

\vspace{-3mm}
Here, $M^k$ is the corresponding random variable of a subset of the input space $\mathcal{M}^k \subset \mathcal{X}$, where each point in $\mathcal{M}^k$ has a unique distance to a fixed anchor point $\mathbf{x}_{k} \in \mathcal{X}$. Together, all $N_p$ partition subsets form the entire input space $\mathcal{X} = \bigcup_{k \in [1, N_p]} \mathcal{M}^k: \bigcap_{k \in [1, N_p]} \mathcal{M}^k = \emptyset$. Further, $f_L^k(.) = d_L(\mathbf{x}_{k}; .)$ is a distance function with respect to the anchor point. We provide a formal definition of unique distance sets $\mathcal{M}^k$, as well as proof for Equation~\ref{eq:minimally-sufficient-statistic-distances} in Appendix~\ref{app:min-statistic-distances}.

Equation~\ref{eq:minimally-sufficient-statistic-distances} is conceptually important as it provides a direct dependency between the learning problem of neural networks and distances in the feature plane. In particular, we note that the compression objective involves minimizing $I(d_H(h_w(\mathbf{x}_{k}); h_w(M^k)); d_X(\mathbf{x}_{k}; M^k))$ which is in direct opposition to preservation constraints that aim to maximize the similarities between the input and feature distances. We further note the maximization objective between the feature distances and the label distribution $I(f_H^k(h_w(M^k)); Y^{k})$. Here, preservation constraints can be beneficial: when additional distances are preserved that contain information related to the label distribution $Y^k$, the term is increased. We arrive at the following observation:


\begin{observation}
\label{obs:distances}
    Preserving distances in the output is beneficial if the preserved distances contribute to the application objective (i.e. contain information of label distribution), and oppose the learning problem otherwise.
\end{observation}

In the following, we analyze the practical scenario of class imbalance to showcase Observation~\ref{obs:distances}.



\begin{figure}
    \begin{center}
        \includegraphics[scale=0.5]{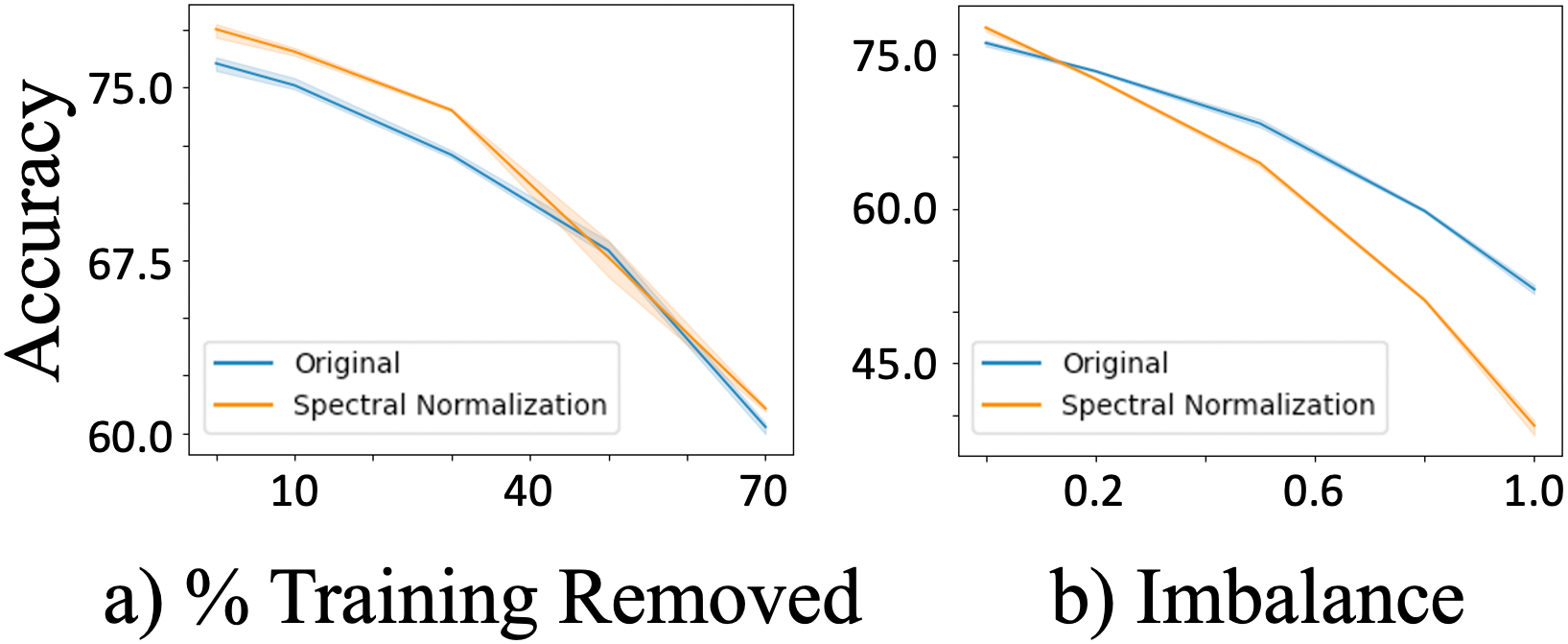}
    \end{center}
    \caption{Classification accuracy with and without distance preservation in the output: a) uniformely removed training and test data (left); b) class imbalance at different severity levels (right). In both graphs, we show the classification accuracy on the y-axis. The x-axis on the left graph represents the percentage of uniformely removed data, on the right the axis represents the fraction of imbalanced classes. The zero point on the x-axis is equivalent for both scenarios and represents the standard CIFAR100 benchmark without imbalance or data removal.}
    \label{fig:acc-analysis}
\end{figure}

\subsection{Distance Preservation under Class Imbalance}

In practice, relevancy information of features is not available and popular options preserve differences between features blindly without regard of application \citep{miyato2018spectral, gulrajani2017improved}. A common example where this characteristic is problematic is class imbalance. Here, information is either over- or under-represented in the training set, resulting in an increase of application-irrelevant data. In this subsection, we investigate the generalization performance when preserving features in the output under different severities of class imbalance. We find that distance constraints result in performance decline under high imbalance severities. In addition to generalization performance, we further investigate the uncertainty estimates under class imbalance in Appendix~\ref{app:uncertainty-surface}.

\vspace{-3mm}

\paragraph{Experimental Setup}
To illustrate class imbalance, we artificially imbalance the CIFAR100 benchmark by removing either training or test samples of a previously balanced class. The portion of classes we artificially imbalance determines the severity of imbalance. For our experiments, we enforce distance preservation through spectral normalization in combination with residual connections \citep{miyato2018spectral}. We choose spectral normalization due to its simplicity and often stronger performance than the double sided gradient penalty \citep{gulrajani2017improved}. We compare other distance-based methods in our benchmark experiments in Section~\ref{sec:results}. Full details on both imbalance method and experimental setup are provided in Appendix~\ref{app:exp-details-analysis}.

\vspace{-3mm}

\paragraph{Accuracy Curves}
We compare the classification accuracy with and without spectral normalization in Figure~\ref{fig:acc-analysis}. In addition to class imbalance, we consider settings where we randomly remove training samples (left graph). We show this setting to determine that the accuracy gain/loss from output feature preservation is dependent on the available information, not the number of samples. Our experiments highlight both advantages and disadvantages of preserving features in the output representation: if the target distribution is sufficiently similar to the input distribution, additional preserved features correlate with the generalization objective and results in a performance increase. This can be seen from the accuracy improvement with spectral normalization under low imbalance severities or when samples are removed randomly. The opposite can be observed where, in contrast to random sample removal (left graph), we explicitly remove training and test samples to imbalance classes (right graph). Here, the training set contains more significant amounts of irrelevant information and spectral normalization significantly decreases the generalization performance.

\vspace{-3mm}
\section{Our Method: TULIP}

\label{sec:methodology}
\subsection{Transitional Feature Preservation}
Within the previous sections we found that feature preservation in the output can oppose the learning objective; an undesirable property for neural networks. To this end, a more prudent strategy involves preserving distance information in a collection of representations instead of a single output. Classic models that implement this property are ensembles \citep{lakshminarayanan2017simple}. Uncertainty is encoded in the collection of ensemble models \emph{without explicit preservation constraints} and the feature distance is then preserved in the difference or ``spread" of the individual representations \citep{malinin2019ensemble}. We formalize the concept within the context of distance preservation: given a set of neural network representations $\{ h_{w_1}(\mathbf{x}), ..., h_{w_N}(\mathbf{x}) \}$ and a transitional function $\Delta h: \mathcal{H}_1 \times \mathcal{H}_2 \times ... \times \mathcal{H}_N \rightarrow \mathcal{V}$, we seek to learn feature mappings that allow a distance $d_V$ within the transitional space that reflects the true distance $d_X$:
\vspace{-3mm}
\begin{equation}
\label{eq:transitional-feature-preservation}
\begin{split}
   d_V(\Delta h(\mathbf{x}_1), \Delta h(\mathbf{x}_2)) = d_X(\mathbf{x}_1, \mathbf{x}_2)
\end{split}
\end{equation}
\vspace{-3mm}

We refer to methods along Equation~\ref{eq:transitional-feature-preservation} as \emph{transitional feature preservation} or TFP in short. While powerful iterative methods such as ensembles implement TFP, their evaluation requires several forward passes and is frequently infeasible due to time or space limitations. In this paper, we implement TFP in single-pass uncertainty estimation by considering a linear combination of intermediate layer representations within the neural network. In particular, we find that the linear combination of intermediate distances $\sum_{l=0}^{L}r_l d_{H_l}(h_{w_l}(\mathbf{x}_1), h_{w_l}(\mathbf{x}_2))$ is distance preserving when the first layer is collapse resistant (i.e. $d_{H_0}(h_{w_0}(\mathbf{x}_1), h_{w_0}(\mathbf{x}_2)) \neq 0)$ for $d_X(\mathbf{x}_1, \mathbf{x}_2) \neq 0$). This requirement is different from Equation~\ref{eq:representational-feature-preservation} as it allows distance contraction or expansion of $d_{H_0}$ with respect to $d_X$, not full preservation. Proposition~\ref{prop:tfp-intermediate} makes the concept more precise. 

\begin{proposition}[Transitional Feature Preservation in Intermediate Representations]
\label{prop:tfp-intermediate}
Consider the neural network mapping $h_w:\mathcal{X}\rightarrow \mathcal{H}$ with the layered architecture $h_w = h_{w_0}\circ h_{w_1}...\circ h_{w_L}$, where the first layer $h_{w_0}$ is collapse resistant with respect to the input space, $d_{H_0}(h_{w_0}(\mathbf{x}_1), h_{w_0}(\mathbf{x}_2)) \neq 0$ for $d_X(\mathbf{x}_1, \mathbf{x}_2) \neq 0$. Then there exists a $C \in \mathbb{R}$ such that

\vspace{-3mm}
\[
\sum_{l=0}^{L}r_l d_{H_l}(h_{w_l}(\mathbf{x}_1), h_{w_l}(\mathbf{x}_2)) = C * d_X(\mathbf{x}_1, \mathbf{x}_2),
\]
where $C=1$ under an appropriate choice of $r_l$. In other words, there exists a linear combination of intermediate representations that is feature preserving in transition - i.e. satisfies Equation~\ref{eq:transitional-feature-preservation}.
\end{proposition}

We provide proof of Proposition~\ref{prop:tfp-intermediate} in Appendix~\ref{app:distortion}. Note that Proposition~\ref{prop:tfp-intermediate} assumes collapse resistance in the first layer. In practice, this can be achieved by enforcing preservation constraints such as spectral normalization on the first layer only, which theoretically comes with the same risks outlined in Section~\ref{sec:theory-overall} (to a significantly lesser degree). Empirically, we found that omitting the constraint entirely does not compromise the quality of the uncertainty estimates suggesting that the first layer rarely collapses data points in practice.

\subsection{Algorithm}
\begin{figure*} [h]
    \begin{center}
        \includegraphics[scale=0.4]{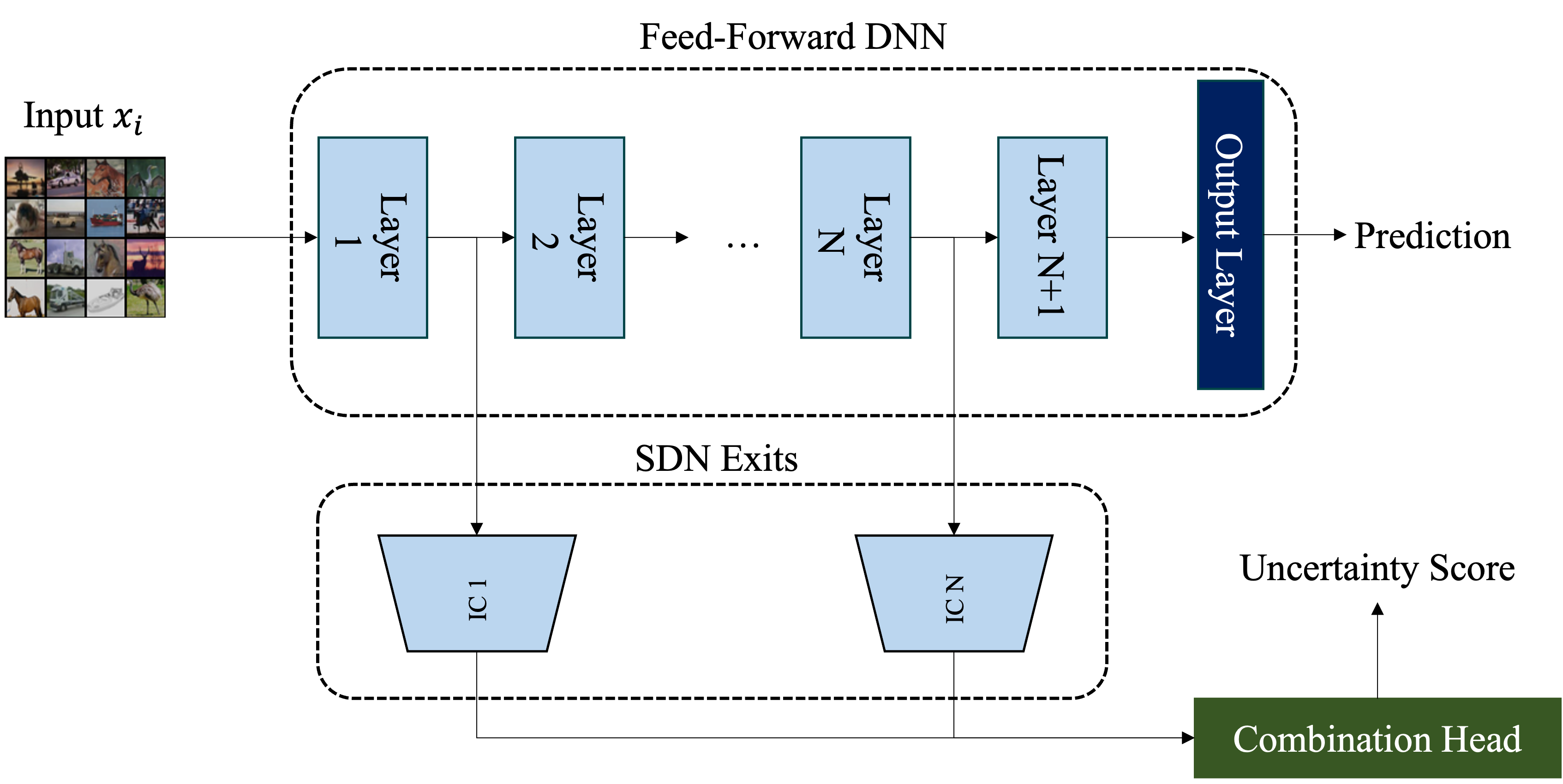}
    \end{center}
    
    \vspace{-3mm}
    \caption{Workflow of our method during inference. The architecture consists of a main network, internal classifiers (IC), as well as a combination head. During inference, the input traverses the main network, as well as the internal classifiers. The prediction is obtained from the main network output, while the uncertainty score is obtained from a combination of internal classifier outputs.}
    \label{fig:turing-workflow}
\end{figure*}

Our method consists of three components: a principal feed-forward network, a constellation of shallow-deep network exits with individual internal classifiers, and a combination head (Figure~\ref{fig:turing-workflow}). During training, the shallow-deep network exits are trained jointly with the feed-forward component, while the combination head is fitted after optimization on a validation set extracted from the training data $\mathcal{X}_{ID}$. We emphasize that our method \emph{does not require out-of-distribution validation samples} as the combination head is fitted with data from the training set. During inference, the input sample traverses both the feed-forward network, as well as the shallow exits. The final output prediction is derived from the main network, while the uncertainty score is derived from a combination of the intermediate output logits. In addition to our description in the main paper, we provide implementation details and algorithm pseudo code in Appendix~\ref{app:summary-method}.

\vspace{-3mm}

\paragraph{Shallow-Deep Network Exits}
Shallow-Deep networks (SDNs) \citep{kaya2019shallow} were originally introduced in the context of computation reduction. A SDN is a modified version of a conventional DNN where additional internal classifiers are placed on intermediate representations to produce preliminary predictions. In our context, we utilize internal classifiers to produce intermediate logits for uncertainty estimation. Formally, given the intermediate layer $l$ of the principal feed-forward network with the latent representation $h^l_{w_l}(\mathbf{x}) \in \mathbb{R}^{M_l} $, an internal classifier is a shallow network $f^l_{w_l}: \mathbb{R}^{M_l} \rightarrow \mathbb{R}^K$ that produces prediction logits for $K$ classes. Regarding positioning, we place internal classifiers uniformly across the feed forward network \cite{kaya2019shallow}.


\vspace{-3mm}

\paragraph{Training Procedure}
We train internal classifiers jointly with the feed-forward network, similar to \cite{kaya2019shallow}. For this purpose, we propose a weighted loss function for each internal classifier, as well as the principal network output. Given the final logits of the principle network $h_w(\mathbf{x})$, an appropriate proper scoring rule (or loss) $L_s$, the SDN Loss is defined as 

\vspace{-3mm}
\begin{equation}
	\label{eq:sdn-loss}
	\begin{split}
		L_{SDN}(\mathbf{x}, y) = p_0 L_s(h_w(\mathbf{x}), y) + \sum_{i=1}^{N_{IC}} p_i L_s(f^i_{w_l}(\mathbf{x}), y).
	\end{split}
\end{equation}

\vspace{-3mm}

where $p_i$ represent individual loss weights, and $N_{IC}$ is the total number of internal classifiers. In our experiments, we found that equal weighting (i.e. $p_k = \frac{1}{N_{IC} + 1})$) produces sufficient results but emphasize that other weighting schemes may be more desirable depending on the application.

\vspace{-3mm}
\paragraph{Combination Head}
While heavily utilizing early layer representations is intuitive, we found that relying on shallow outputs exclusively results in poor uncertainty scores. Specifically, in addition to increased domain awareness, early intermediate outputs contain significant amounts of application-irrelevant information that results in unreliable uncertainty estimates. 
For this purpose, we propose a fusion scheme by first calculating individual uncertainty scores and subsequently combining them by a weighted scaled sum.

\vspace{-6mm}
\begin{equation}
	\label{eq:sdn-uncertainty}
	\begin{split}
		u_{final}(\mathbf{x}) = \frac{1}{\sum_{i=1}^{N_{IC}} r_i} \sum_{i=1}^{N_{IC}} r_i u_s(f^i_{w_i}(\mathbf{x})).
	\end{split}
\end{equation}
In Equation~\ref{eq:sdn-uncertainty}, $u_s$ represents the individual uncertainty score, and $r_i$ the score weights. While principally any uncertainty score can be used, we replace the output layer of the internal classifiers with approximate gaussian processes and calculate uncertainty similar to \citep{liu2020simple}. Our choice is based on simplicity and the strong performance of approximate Gaussian processes. We call our resulting method \emph{Transitional Uncertainty with Layered Intermediate Predictions} or \texttt{TULIP} in short.

\vspace{-3mm}

\begin{table*}
\centering
\caption{OOD detection and classification accuracy on the CIFAR10 dataset.}
\label{tab:cifar10-ood-rn}
\tiny
\begin{tabularx}{\textwidth}{l X c c c c c c c}
\hline
              &           &  & &   &     & \multicolumn{3}{c}{\centering OOD AUROC} \\
              Architecture & Algorithms & Runtime & Space Complexity & Preservation Constraint & Accuracy & CIFAR10-C & CIFAR100-C & SVHN \\
		\hline
		\hline
        \multirow{5}{*}{ResNet-50}
		& Ensemble of 5 \citep{lakshminarayanan2017simple}     & 5x & 5x & No Constraint & 96.327 ± 0.045       & 0.725 ± 0.002      & 0.914 ± 0.003      & 0.943 ± 0.008      \\
        & DNN               & \multirow{2}{*}{1x} & \multirow{2}{*}{1x} & \multirow{2}{*}{No Constraint}  & \bf{\multirow{2}{*}{95.533 ± 0.080}}       & 0.715 ± 0.007      & 0.903 ± 0.000      & 0.927 ± 0.015        \\
    	& Energy-Based \citep{liu2021energybased}      &  & & &     & 0.688 ± 0.016      & 0.879 ± 0.003      & 0.881 ± 0.006       \\
		& DUN\citep{antoran2020depth}     & 1x & 1.5x & Div. Loss & 94.675 ± 0.045       & 0.713 ± 0.003      & 0.881 ± 0.006      & 0.880 ± 0.008      \\
		& Early Exit\citep{qendro2021early}     & 1x & 1.5x & Div. Loss & 93.510 ± 0.100       & 0.714 ± 0.005      & 0.881 ± 0.007      & 0.870 ± 0.009      \\
    	& SNGP\citep{liu2020simple} & 1x & 1x & SN & 95.033 ± 0.076       & 0.721 ± 0.007      & 0.928 ± 0.005      & \bf{0.976 ± 0.003}      \\
		& DUQ\citep{van2020duq}     & 1x & 1x & GP & 88.867 ± 0.211       & 0.618 ± 0.003      & 0.824 ± 0.008      & 0.829 ± 0.016      \\
        \cline{2-9}
    	& \texttt{TULIP}     & 1x & 1.5x &  No Constraint   & 94.880 ± 0.324       & \bf{0.738 ± 0.006}      & \bf{0.936 ± 0.003}      & 0.946 ± 0.012        \\
		\hline
		\hline
        \multirow{5}{*}{ResNet-101}
		& Ensemble of 5 \citep{lakshminarayanan2017simple}     & 5x & 5x & No Constraint & 96.237 ± 0.057       & 0.706 ± 0.001      & 0.908 ± 0.000      & 0.939 ± 0.004      \\
        & DNN              & \multirow{2}{*}{1x} & \multirow{2}{*}{1x} & \multirow{2}{*}{No Constraint}  & \bf{\multirow{2}{*}{95.837 ± 0.103}}       & 0.690 ± 0.006      & 0.894 ± 0.001      & 0.922 ± 0.012        \\
    	& Energy-Based \citep{liu2021energybased}     & & &  &        & 0.715 ± 0.006      & 0.849 ± 0.009      & 0.890 ± 0.034       \\
     
		& DUN\citep{antoran2020depth}     & 1x & 1.5x & Div. Loss & 95.645 ± 0.075       & 0.711 ± 0.003      & 0.889 ± 0.005      & 0.917 ± 0.014  \\
		& Early Exit\citep{qendro2021early}     & 1x & 1.5x & Div. Loss & 94.255 ± 0.085       & 0.706 ± 0.001      & 0.875 ± 0.003      & 0.867 ± 0.045      \\
  
    	& SNGP\citep{liu2020simple} & 1x & 1x & SN & 91.907 ± 0.183       & 0.636 ± 0.009      & 0.912 ± 0.015      & 0.906 ± 0.018      \\
		& DUQ\citep{van2020duq}     & 1x & 1x & GP & 89.427 ± 0.315       & 0.620 ± 0.003      & 0.833 ± 0.004      & 0.830 ± 0.004      \\
        \cline{2-9}
    	& \texttt{TULIP}  & 1x & 1.5x &  No Constraint   & 94.257 ± 0.349       & \bf{0.722 ± 0.006}    & \bf{0.937 ± 0.003}      & \bf{0.938 ± 0.004}        \\
     \hline
	 \hline
        \multirow{5}{*}{ResNet-152}
		& Ensemble of 5 \citep{lakshminarayanan2017simple}     & 5x & 5x & No Constraint & 96.330 ± 0.160       & 0.706 ± 0.001      & 0.910 ± 0.000      & 0.945 ± 0.000      \\
        & DNN               & \multirow{2}{*}{1x} & \multirow{2}{*}{1x} & \multirow{2}{*}{No Constraint} & \bf{\multirow{2}{*}{95.877 ± 0.097}}       & 0.690 ± 0.006      & 0.891 ± 0.000      & 0.928 ± 0.005        \\
    	& Energy-Based \citep{liu2021energybased}    & & &  &        & 0.665 ± 0.008      & 0.849 ± 0.004      & 0.905 ± 0.018      \\
     
		& DUN\citep{antoran2020depth}     & 1x & 1.5x & Div. Loss & 95.850 ± 0.190       & 0.693 ± 0.006      & 0.884 ± 0.003      & 0.883 ± 0.019  \\
		& Early Exit\citep{qendro2021early}     & 1x & 1.5x & Div. Loss & 94.340 ± 0.085       & 0.679 ± 0.000      & 0.836 ± 0.001      & 0.796 ± 0.000      \\
  
    	& SNGP\citep{liu2020simple} & 1x & 1x & SN & 90.510 ± 0.814       & 0.636 ± 0.009      & 0.899 ± 0.016      & 0.847 ± 0.020      \\
		& DUQ\citep{van2020duq}     & 1x & 1x & GP & 91.263 ± 0.185       & 0.623 ± 0.002      & 0.731 ± 0.059      & 0.842 ± 0.012      \\
        \cline{2-9}
    	& \texttt{TULIP}   & 1x & 1.5x &  No Constraint  & 94.213 ± 0.777       & \bf{0.722 ± 0.006}    & \bf{0.927 ± 0.004}      & \bf{0.945 ± 0.004}        \\
\end{tabularx}

\end{table*}

\paragraph{Fitting the Combination Head}
\label{sec:fitting-combination-head}
Fitting the weights $r_i$ in Equation~\ref{eq:sdn-uncertainty} is more involved. While several methods exist to fit uncertainty parameters \citep{lee2023probing, lee2018simple, liang2017enhancing}, they require either a) access to a small set of $\mathcal{X}_{OOD}$, and/or b) access to all labels in $\mathcal{X}_{ID}$. In this paper, we assume neither. Our fitting algorithm requires two steps. First, we derive proxy labels from the small validation set. Second, we derive the weight parameters by formulating a binary classification problem, using the individual uncertainty scores $u_s(f^i_{w_i}(\mathbf{x}))$, and proxy labels $s(\mathbf{x})$ derived from the disagreement in between different SDN exits. In Appendix~\ref{app:disagreement-weight-tuning}, we study the proxy labels in detail and compare disagreement in SDN exits with ensembles. Note, that our algorithm assumes that the validation set represents a small amount of \emph{unlabeled} samples originating from $\mathcal{X}_{ID}$ and does \emph{not} require out-of-distribution samples or additional data of any kind. We define our proxy labels through disagreement in the form of prediction switches between internal classifiers. Given a validation sample $\mathbf{x}_{val}$, we define a prediction switch as

\vspace{-3mm}

\begin{equation}
	\label{eq:sdn-subgroup-classification}
	\begin{split}
		s(\mathbf{x}_{val}) = bool(\sum_{i=2}^{N_{IC}} \mathbf{1}_{f^i_{w_i} != f^{i-1}_{w_{i-1}}} < N_{s})
	\end{split}
\end{equation}

\vspace{-3mm}

\begin{table*}[!ht]
\centering
\caption{OOD detection and classification accuracy on the CIFAR100 dataset.}
\tiny
\label{tab:cifar100-ood-rn}
\begin{tabularx}{\textwidth}{l X c c c c c}
\hline
              &           &     &     & \multicolumn{3}{c}{\centering OOD AUROC} \\
              Architecture & Algorithms & Preservation Constraint & Accuracy & CIFAR10-C & CIFAR100-C & SVHN \\
		\hline
		\hline
        \multirow{5}{*}{ResNet-50}
		& Ensemble of 5 \citep{lakshminarayanan2017simple}     & No Constraint & 79.475 ± 0.265       & 0.828 ± 0.003      & 0.680 ± 0.002      & 0.857 ± 0.009      \\
        & DNN              & \multirow{2}{*}{No Constraint} & \multirow{2}{*}{77.623 ± 0.276}       & 0.832 ± 0.003      & 0.680 ± 0.004     & 0.859 ± 0.013       \\
        & Energy-Based \citep{liu2021energybased}    &  &  & 0.829 ± 0.008      & 0.682 ± 0.006      & 0.815 ± 0.050       \\
     
		& DUN\citep{antoran2020depth}     & Div. Loss & 75.760 ± 0.440       & 0.774 ± 0.002      & 0.648 ± 0.000      & 0.789 ± 0.009  \\
		& Early Exit\citep{qendro2021early}     & Div. Loss & 70.715 ± 0.065       & 0.823 ± 0.004      & 0.883 ± 0.009      & 0.815 ± 0.014      \\
  
    	& SNGP\citep{liu2020simple} & SN & 75.083 ± 0.889       & 0.821 ± 0.010      & 0.707 ± 0.005      & 0.900 ± 0.008     \\
		& DUQ\citep{van2020duq}     & GP & -       & -      & -      & -     \\
        \cline{2-7}
    	& \texttt{TULIP}  &  No Constraint & \bf{78.437 ± 0.223}       & \bf{0.868 ± 0.002}      & \bf{0.738 ± 0.002}      & \bf{0.955 ± 0.008}        \\
		\hline
		\hline

        \multirow{5}{*}{ResNet-101}
		& Ensemble of 5 \citep{lakshminarayanan2017simple}     & No Constraint & 79.545 ± 0.015       & 0.829 ± 0.002      & 0.680 ± 0.002      & 0.849 ± 0.006      \\
        & DNN             & \multirow{2}{*}{No Constraint}  & \multirow{2}{*}{77.257 ± 0.285}       & 0.834 ± 0.003      & 0.671 ± 0.005      & 0.851 ± 0.024       \\
        & Energy-Based \citep{liu2021energybased}   &   &   & 0.836 ± 0.004      & 0.674 ± 0.007      & 0.856 ± 0.042       \\
     
		& DUN\citep{antoran2020depth}     & Div. Loss & 77.680 ± 0.013       & 0.780 ± 0.001      & 0.647 ± 0.000      & 0.783 ± 0.002  \\
		& Early Exit\citep{qendro2021early}     & Div. Loss & 71.215 ± 0.045       & 0.810 ± 0.001      & 0.670 ± 0.002      & 0.828 ± 0.005      \\
  
        & SNGP\citep{liu2020simple} & SN & 74.380 ± 1.978       & 0.816 ± 0.028      & 0.674 ± 0.031     & 0.906 ± 0.027     \\
        & DUQ\citep{van2020duq}     & GP & -       & -      & -      & -     \\
        \cline{2-7}
        & \texttt{TULIP}   &  No Constraint & \bf{78.550 ± 0.213}      & \bf{0.863 ± 0.004}   &  \bf{0.730 ± 0.002}     & \bf{0.959 ± 0.001}        \\
     \hline
		\hline

        \multirow{5}{*}{ResNet-152}
		& Ensemble of 5 \citep{lakshminarayanan2017simple}     & No Constraint & 78.487 ± 0.133       & 0.725 ± 0.002      & 0.680 ± 0.002      & 0.822 ± 0.002      \\
        & DNN             & \multirow{2}{*}{No Constraint}  & \multirow{2}{*}{78.160 ± 0.242}       & 0.830 ± 0.002     & 0.674 ± 0.002      & 0.851 ± 0.010        \\
    	& Energy-Based \citep{liu2021energybased} & &  & 0.828 ± 0.002      & 0.674 ± 0.002      & 0.852 ± 0.009       \\
     
		& DUN\citep{antoran2020depth}     & Div. Loss & \bf{78.845 ± 0.025}       & 0.721 ± 0.004      & 0.637 ± 0.003      & 0.796 ± 0.009  \\
		& Early Exit\citep{qendro2021early}     & Div. Loss & 72.490 ± 0.005       & 0.816 ± 0.007      & 0.675 ± 0.006      & 0.828 ± 0.005      \\
  
    	& SNGP\citep{liu2020simple} & SN & 74.077 ± 2.631       & 0.822 ± 0.028      & 0.659 ± 0.039      & 0.889 ± 0.015    \\
    	& DUQ\citep{van2020duq}     & GP & -       & -      & -      & -     \\
        \cline{2-7}
    	& \texttt{TULIP} &  No Constraint & \bf{78.877 ± 0.311}       & \bf{0.857 ± 0.010}    & \bf{0.715 ± 0.010}      & \bf{0.926 ± 0.031}        \\
\end{tabularx}
\end{table*}

where $f^i_{w_i}$ are abbreviations for the internal classifier predictions $pred(f^i_{w_i}(\mathbf{x}_{val})$, and $ \mathbf{1}_{f^i_{w_i} != f^{i-1}_{w_{i-1}}}$ is a binary variable reducing to one if two subsequent classifier predictions differ or zero otherwise. $N_s$ represents the number of switches to determine a positive sample and is an integer in the range $[1, N_{IC}]$. Our choice regarding the disagreement label is based on simplicity. By evaluating prediction switches, we reduce the tuning process to a binary classification problem allowing a partition of the validation set into coarse high-, and one low-uncertainty subgroups. Specifically, we classify the sample $\mathbf{x}_{val}$ as high-uncertainty if $s(\mathbf{x}_{val})$ amounts to one and as low-uncertainty otherwise. Subsequently, we derive the weighting parameters through logistic regression, where we map the individual uncertainty scores to the $u_s(f^i_{w_i}(\mathbf{x}_{val}))$ to the corresponding subgroup $s(\mathbf{x}_{val})$:

\vspace{-3mm}

\begin{equation}
	\label{eq:sdn-subgroup-logregression}
	\begin{split}
		r_1, ..., r_{N_{IC}} = LR(\{ s(\mathbf{x}_{i}), \mathbf{v}_i \}_{i=1}^{N_{val}}) 
	\end{split}
\end{equation}

\vspace{-2mm}

In our notation, $LR$ is an abbreviation for logistic regression, and $\mathbf{v}_i$  are the individual uncertainty scores $[u_s(f^1_{w_1}(\mathbf{x}_i)), ..., u_s(f^{N_{IC}}_{w_{N_{IC}}}(\mathbf{x}_i))]$ bundled into a single vector.





\section{Related Work}

Our work most closely relates with estimating uncertainty in a single forward pass. Several initial studies in single-pass uncertainty estimation were considered within the context of online settings or regression tasks. Here, notable methods include quantile regression \citep{quantileregression}, conformal prediction \citep{shafer2008tutorial}, or direct variance prediction \citep{directvar}. The initial concepts were followed by a large body of uncertainty estimation methods including but not limited to replacing the loss function \citep{malinin2018predictive, hein2019relu, sensoy2018evidential}, the output layer \citep{liu2021energybased, padhy2020revisiting, bendale2016towards, macedo2022enhanced, tagasovska2019single, gast2018lightweight},  alternative gradient representations \citep{kwon2020backpropagated, prabhushankarintrospective, lee2023probing}, or interval arithmetic \citep{oala2020interval}. Notably, \cite{gast2018lightweight} modify the network to produce probabilistic outputs instead of point estimates and \cite{oala2020interval} propagate error intervals directly through the neural network. While several methods are promising, they do not explicitly consider information preservation within the representations of the network and are susceptible to collapsing features to single entities. As a result, recent methods consider feature preservation within the output as a vital component for reliable uncertainty scores \citep{liu2020simple, van2020duq, van2021due, mukhoti2023deep}. Our work complements these approaches by considering feature preservation with intermediate representations without explicit constraints. Further, additional approaches exist for uncertainty estimation \citep{lee2023probing, lee2018simple, guo2017calibration, liang2017enhancing} where several explore intermediate representations. However, they assume access to out-of-distribution validation samples and/or a fully labeled training set. Our work is complementary by investigating feature preservation without access to a out-of-distribution validation set and does not assume that all samples are labeled in training. Additionally, there exists several recent studies to bayesian models \citep{evidentialturing}. While theoretically appealing, the framework relies on a bayesian model which requires several iterative forward passes for evaluation. Finally, our work closely relates to several approaches utilizing early-exit neural networks with the application of uncertainty estimation \cite{qendro2021early,antoran2020depth}. While initial results are promising, these approaches utilize a diversity loss term which can be interpreted as a feature preservation constraint along Equation~\ref{eq:representational-feature-preservation}. Our work is complementary by utilizing early-exit neural networks without feature preservation constraints. We compare against two other early-exit approaches in our experiments. In Appendix~\ref{app:distance-preserving}, we provide further related work on iterative uncertainty estimation, as well as distance-preserving neural networks. 

\section{Benchmark Experiments}

\label{sec:results}

\begin{table*}
\centering
\caption{AUROC and classification accuracy on on the organ\{C, A, S\} datasets.}
\label{tab:organ-ood-rn}
\tiny
\begin{tabularx}{\textwidth}{l X c c c c c}
\hline
              &           &  &        & \multicolumn{3}{c}{\centering AUROC} \\
              Dataset & Algorithms & Preservation Constraint & Accuracy & organA & organC & organS \\
		\hline
		\hline
        \multirow{4}{*}{organA}
        & DNN               & \multirow{2}{*}{No Constraint} & \multirow{2}{*}{94.602 ± 0.388}       & -      & 0.906 ± 0.004      & 0.850 ± 0.007        \\
    	& Energy-Based \citep{liu2021energybased}      & &     & -      & 0.887 ± 0.005      & 0.841 ± 0.008       \\
     
		& DUN\citep{antoran2020depth}           & Div. Loss & 93.784 ± 0.106       & -      & 0.850 ± 0.008      & 0.747 ± 0.006  \\
		& Early Exit\citep{qendro2021early}     & Div. Loss & 92.009 ± 0.213       & -      & 0.894 ± 0.004      & 0.732 ± 0.002      \\
  
    	& SNGP\citep{liu2020simple} & SN & 93.906 ± 0.297       & -      & \bf{0.907 ± 0.010}      & 0.857 ± 0.006      \\
        \cline{2-7}
    	& \texttt{TULIP}    &  No Constraint  & \bf{95.254 ± 0.191}       & -      & \bf{0.915 ± 0.003}      & \bf{0.869 ± 0.003}        \\
		\hline
		\hline
        \multirow{4}{*}{organC}
        & DNN               & \multirow{2}{*}{No Constraint} & \bf{\multirow{2}{*}{92.106 ± 0.176}}       & 0.884 ± 0.001      & -      & 0.780 ± 0.005        \\
    	& Energy-Based \citep{liu2021energybased}  &   &     & 0.857 ± 0.001      & -      & 0.751 ± 0.006       \\
     
		& DUN\citep{antoran2020depth}           & Div. Loss & 91.231 ± 0.000       & 0.874 ± 0.002      & -      & \bf{0.801 ± 0.001}  \\
		& Early Exit\citep{qendro2021early}     & Div. Loss & 90.826 ± 0.236       & 0.867 ± 0.000      & -      & 0.791 ± 0.001      \\
  
    	& SNGP\citep{liu2020simple} & SN & 90.941 ± 0.530       & 0.849 ± 0.008      & -      & 0.765 ± 0.007      \\
        \cline{2-7}
    	& \texttt{TULIP}     &  No Constraint  & \bf{92.122 ± 0.227}       & \bf{0.894 ± 0.003}      & -      & 0.794 ± 0.003        \\
		\hline
		\hline
        \multirow{4}{*}{organS}
        & DNN               & \multirow{2}{*}{No Constraint} & \multirow{2}{*}{80.258 ± 0.299}      & 0.754 ± 0.010      & 0.815 ± 0.003      & -        \\
    	& Energy-Based \citep{liu2021energybased}     & &     & 0.733 ± 0.012      & 0.775 ± 0.005      & -       \\
     
		& DUN\citep{antoran2020depth}           & Div. Loss & \bf{81.657 ± 0.198}       & 0.751 ± 0.003      & \bf{0.824 ± 0.005}      & -  \\
		& Early Exit\citep{qendro2021early}     & Div. Loss & 78.888 ± 0.000       & 0.747 ± 0.001      & 0.815 ± 0.003      & -      \\
  
    	& SNGP\citep{liu2020simple} & SN & 79.918 ± 0.280       & 0.707 ± 0.009      & 0.790 ± 0.010      & -      \\
        \cline{2-7}
    	& \texttt{TULIP}      &  No Constraint  & 80.002 ± 0.126       & \bf{0.778 ± 0.001}      & \bf{0.822 ± 0.003}      & -        \\
\end{tabularx}

\end{table*} 

\subsection{CIFAR10 and CIFAR100}
We start with standardized benchmarks in out-of-distribution (OOD) detection. The following combinations are evaluated: CIFAR10 vs. CIFAR10-C/CIFAR100-C/SVHN and CIFAR100 vs. CIFAR10-C/CIFAR100-C/SVHN \citep{cifar, svhn, cifar10c}. In addition to a standard DNN, we compare against five single-pass uncertainty baselines that do not require additional OOD data: the energy-based model \citep{liu2021energybased}, DUN \cite{antoran2020depth}, early-exit ensembles \cite{qendro2021early}, DUQ \citep{van2020duq}, and SNGP \citep{liu2020simple}. We choose these three methods because 1) their strong empirical performance and 2) they utilize four popular methods for feature preservation in the network output. DUQ preserves features with a double sided gradient penalty (GP) \citep{gulrajani2017improved}, SNGP implements spectral normalization (SN) \citep{miyato2018spectral}, DUN and early-exit ensembles utilize a an additional diversity loss term (Div. loss) \cite{qendro2021early, antoran2020depth}, and the energy based model relies on the softmax density without regularization (No constraint). To investigate robustness with respect to network complexity, we consider three architectures with residual connections and varying depth: ResNet architectures \citep{he2016deep} with 50, 101, and 152 layers. We restrict our experiments to these architectures as SNGP requires residual connections for its functionality. For fine-tuning the uncertainty weights, we partition 10\% of training the samples and remove the labels to perform the unsupervised fitting algorithm. In Table~\ref{tab:cifar10-ood-rn} and Table~\ref{tab:cifar100-ood-rn}, we report the AUROC scores for training on CIFAR10 and CIFAR100 respectively. When evaluating the corruption datasets, CIFAR10-C and CIFAR100-C, we average all corruption types and intensities. Further details on implementation and feature preservation can be found in Appendix~\ref{app:ood-c10/100} and \ref{app:com-method-details}. In addition, we investigate calibration, runtime, and imbalanced settings in Appendix~\ref{app:calibration-experiments}, and~\ref{app:imbalanced-ood}.

Our method outperforms the other single-pass methods despite having access to less training annotations. This holds true over varying architectures and training datasets. In particular, methods trained with spectral normalization achieve lower accuracy with increasing network depth. For instance, the accuracy of SNGP reduces nearly three percent when extending ResNet-50 to ResNet-101. We relate this observation to the scaling of the Lipschitz bounds. Specifically, the lower and upper bounds scale exponentially with the number of layers and are tighter for shallow models while looser for deeper ones. Hence, the preservation constraint is weaker for deeper models and fails to maintain the distance between data-points. The remaining methods deploy a different feature preservation strategy and are thus agnostic to this effect. Further, DUQ did not converge on CIFAR100 due to training instabilities. These arise when the class centroids get noisy from increasing class and data complexity. Finally, we acknowledge the runtime and space complexity. While our method is favorable in terms of performance adding internal classifiers does increase the space complexity of the model. However, the effect does not scale with the network depth and is significantly less than ensembles which is 2x minimum.

\begin{table*}
\tiny
\centering
\caption{OOD detection and classification accuracy on the ImageNet dataset.}
\label{tab:imgnet-rn}
\begin{tabularx}{\textwidth}{X c c c}
\hline
              Algorithms & Accuracy & ECE & ImageNet-C \\
		\hline
		\hline
        DNN                & \multirow{3}{*}{72.617 ± 0.019}       & \multirow{2}{*}{1.138 ± 0.005} & 0.867 ± 0.006    \\
    	Energy-Based\citep{liu2021energybased}       &     &                    & 0.876 ± 0.004      \\
    	MC-Dropout\citep{gal2016dropout}         &     & 0.837 ± 0.001      & 0.877 ± 0.000     \\
        \hline
		\texttt{TULIP}    & 72.617 ± 0.019   & \bf{0.678 ± 0.001} & \bf{0.885 ± 0.000}   \\
\end{tabularx}

\end{table*}

\subsection{Medical Modalities}
\label{sec:medical-modalities}
Further limitations of current single-pass methods include their strong modification the baseline algorithm. This renders them difficult to scale to different data modalities. For this purpose, we consider a medical setting where the training data contains different information as the test set. We benchmark \texttt{TULIP} on three CT scan datasets from \cite{medmnistv2}. All three datasets contain CT scans of the same eleven body organs and are named after the three planes (axial, coronal, and saggittal) in which the data was collected. In our experiments, we train on one plane and perform misclassification detection on the combined test set of the original plane and an additional plane. We report AUROC and accuracy on the in-domain test set in Table~\ref{tab:organ-ood-rn}. Each row shows a different training set, while each column refers to the test set that is combined with the in-distribution test set. All experiments are performed with a ResNet-50 architecture and we use the same hyperparameters as in our previous experiments. 

\texttt{TULIP} matches or outperforms competing methods. The observation is important as it shows that our modifications scale to complex data modalities which is not the case for several other single-pass methods. In particular, algorithms with strong architectural modifications such as SNGP does not perform well in the medical modality. The result is expected due to imbalance in the dataset. In particular, SNGP is challenged when the information within the training set does not correlate well with the test set and supports our usage of TFP in \texttt{TULIP}. These results are complementary and support conclusions from existing studies \citep{postels22apracticality}.

\subsection{ImageNet}
\label{app:imgnet-results}
In this section, we investigate how \texttt{TULIP} adapts when internal classifiers are not trained together with the baseline architecture. A further limitation of several single-pass uncertainty estimators is that their implementation requires a significant change to the existing architecture. In several applications the modification is non-trivial, especially in large-scale benchmarks. For this purpose, we consider settings where we do not modify the baseline architecture, allowing the same generalization performance to a standard DNN. In our experiments, we place individual internal classifiers on top of a ResNet-101 architecture fully trained on imagenet and compare against other methods that require no change to the existing architecture. In our experiments, this is the energy-based model \citep{liu2021energybased} as a single-pass example and MC-Dropout \citep{gal2016dropout} as an iterative example. For MC-Dropout, we perform inference in a single forward pass without any dropout layers rendering the same behavior as the baseline DNN. To derive uncertainty estimates, we perform several forward passes with dropout enabled making it an iterative method. We show our results in Table~\ref{tab:imgnet-rn} and provide experimental details in Appendix~\ref{app:imgnet}. From our results, we see that \texttt{TULIP} outperforms several other methods on large scale benchmarks. We interpret these results as a further testament to the feature extraction capabilities of intermediate representations. In particular, our results show that the intermediate representations do not need to be explicitly trained to extract additional information but can be used post-hoc. We further note that \texttt{TULIP} can scale to large scale benchmarks which is not the case for other single-pass methods that need to be trained from scratch. 
\vspace{-3mm}
\section{Discussion and Limitations}

A central observation we made in this work is that \emph{enforcing feature preservation by constraining model representations can be harmful to the model performance}. We highlight application relevance as a key requirement for effective representation constraints. In practice, the characteristic can be undesirable as training distributions can severely differ from deployment. We propose single-pass transitional feature preservation through intermediate representations to address these disadvantages. While our approach is effective, we do not claim that our improvements solve feature preservation in uncertainty estimation entirely. In particular, we propose one instance of TFP through SDNs that comes with its own set of limitations: similar to iterative methods, the success depends on the amount of source representations in $\Delta h$ to preserve features. For this purpose, SDNs are less effective on shallow architectures with fewer intermediate options to extract from. Further, we acknowledge that implementing internal classifiers comes with a large set of hyperparameters increasing space complexity. In our implementation, we use standard settings that show promising performance without significantly impacting the space complexity. However, other applications may require more elaborate hyperparameter explorations and additional internal classifiers. For this reason, we see reducing the space complexity of internal classifiers or the development of efficient hyperparameter tuners as promising research directions. Finally, we chose the combination of SDNs with Gaussian Processes out of simplicity and the strong empirical performance. However, key novelties of this paper (SDNs and singe-pass TFP) are not limited to one uncertainty score (Gaussian Processes) and are building blocks for single-pass uncertainty methods. We encourage researchers to implement different combinations of single-pass TFP and uncertainty scores to advance the field of single-pass uncertainty estimation.



\section*{Impact Statement}
This paper presents work whose goal is to advance the field of Machine Learning. There are many potential societal consequences of our work, none which we feel must be specifically highlighted here.



\bibliography{example_paper}
\bibliographystyle{icml2024}

\newpage
\appendix
\onecolumn
\section{Proofs}

\subsection{Mutual Information of Distances}
\label{app:min-statistic-distances}
In this section, we discuss connecting the learning problem in Equation~\ref{eq:minimal-sufficient-statistic} with distances in the feature plane. To establish a dependency between Equation~\ref{eq:minimal-sufficient-statistic} and feature distances, we first define sets where the distance $d_X$ is bijective. Mathematically, this is equivalent to restricting sets to unique distance values; sets we define as unique distance sets. Definition~\ref{def:unique-distance-set} formalizes the concept.

\begin{definition}[Unique Distance Set and Partition]
\label{def:unique-distance-set}
Consider the metric space $(\mathcal{X}, d_X)$ with a corresponding random variable $X \sim p_X$ describing the input distribution with density $p_X$. We define a unique distance set $\mathcal{M} \subset \mathcal{X}$ as a set in $\mathcal{X}$ possessing unique distances with respect to $d_X$ and an arbitrary but fixed anchor point $\mathbf{x}_a \in \mathcal{X}$. 

\[
\mathcal{M} = \{ \mathbf{x}_i, \mathbf{x}_j, \mathbf{x}_a \in \mathcal{X}: d_X(\mathbf{x}_a, \mathbf{x}_j) \neq d_X(\mathbf{x}_a, \mathbf{x}_i),~i\neq j \neq a \}
\]

Further, we define a partition $\mathcal{X} = \bigcup_{k \in [1, N_p]} \mathcal{M}^k: \bigcap_{k \in [1, N_p]} \mathcal{M}^k = \emptyset$ over unique distance sets $\mathcal{M}^k$ as a unique partition set. Equivalently, we define $M \sim p_M$ and $M^k \sim p_{M^k}$ as the corresponding random variables with their respective probability densities.
\end{definition}

With the help of unique distance sets, we can formulate the proof for Equation~\ref{eq:minimally-sufficient-statistic-distances}.

\begin{proof}

We first formulate Equation~\ref{eq:minimal-sufficient-statistic} as an optimization objective over unique distance sets. Since the optimization over each individual set can be viewed as a separate learning task, we can rewrite the objective as a summary of the mutual information over individual distance sets:

\begin{equation}
\label{eq:mutual-info-partitioned}
\begin{split}
   h^*(X) = \argmin_{h_w: \{ I(h_w(M^k); Y^k) = I(M^k; Y^k),~k \in [1, N_p]\} } \sum_{k} I(h_w(M^k); M^k)
\end{split}
\end{equation}

Given the anchor point $\mathbf{x}_k$ for $\mathcal{M}^k$, $f_X(\mathbf{x}) = d_X(\mathbf{x}_k; \mathbf{x})$ represents an injection within the individual subset $\mathcal{M}^k$. The characteristic is relevant, as we utilize the transformation invariance property of the mutual information. Assuming $h_w$ preserves the unique distance property according to Equation~\ref{eq:representational-feature-preservation}, we rewrite Equation~\ref{eq:mutual-info-partitioned} as

\begin{equation}
\label{eq:minimal-sufficient-statistic-distances}
\begin{split}
   h^*(X) &= \argmin_{h_w: \{ I(h_w(M^k); Y^k) = I(M^k; Y^k),~k \in [1, N]\} } \sum_{k} I(h_w(M^k); M^k)\\
   &=\argmin_{h_w: \{ I(d_H(h_w(\mathbf{x}_{k}); h_w(M^k)); Y^{k}) = I(M^k; Y^{k}),~k \in [1, N]\} } \sum_{k} I(d_H(h_w(\mathbf{x}_{k}); h_w(M^k));d_X(\mathbf{x}_{k}; M^k))\\
   &=\argmin_{h_w: \{ I(f_H^k(h_w(M^k)); Y^{k}) = I(M^k; Y^{k}),~k \in [1, N_p]\} } \sum_{k} I(f_H^k(h_w(M^k)); f_X^k(M^k)).
\end{split}
\end{equation}
\end{proof}

\subsection{Transitional Feature Preservation of Intermediate Representations}
\label{app:distortion}
In this section, we discuss the proof of Proposition~\ref{prop:tfp-intermediate}. 

\begin{proof}
To prove Proposition~\ref{prop:tfp-intermediate}, we utilize concepts of metric distortion from metric embedding theory \citep{abraham2006advances, chennuru2018measures}. Specifically, neural network layers can be characterized by the distortion they introduce to the input space. We define the network distortion coefficient of a given layer $l$ as

\begin{equation}
\label{eq:network-distortion}
\rho_l(\mathbf{x}_1, \mathbf{x}_2) = \begin{cases}
\frac{d_{H_l}(h_{w_l}(\mathbf{x}_1), h_{w_l}(\mathbf{x}_2))}{d_{H_{l-1}}(h_{w_{l-1}}(\mathbf{x}_1), h_{w_{l-1}}(\mathbf{x}_2))},& \text{if } d_{H_{l-1}}(h_{w_{l-1}}(\mathbf{x}_1), h_{w_{l-1}}(\mathbf{x}_2)) \neq 0\\
1, & \text{otherwise}
\end{cases}.
\end{equation}

The interpretation of Equation~\ref{eq:network-distortion} is simple. If the previous layer does not collapse the input distances ($d_{H_{l-1}}(h_{w_{l-1}}(\mathbf{x}_1), h_{w_{l-1}}(\mathbf{x}_2)) \neq 0$), the ratio between both distances characterizes the distance distortion of layer $l$. Specifically, we have \emph{distance contraction} if $\rho_l(\mathbf{x}_1, \mathbf{x}_2)$ is less than one. Here, zero represents the corner case when the layer collapses the input to a single point ($d_{H_l}(h_{w_l}(\mathbf{x}_1), h_{w_l}(\mathbf{x}_2)) = 0$). Further, we have \emph{distance expansion} when $\rho_l(\mathbf{x}_1, \mathbf{x}_2)$ is greater than one, i.e. when the layer increases the distance between the points. Finally, we have perfect distance preservation in the case of $\rho_l(\mathbf{x}_1, \mathbf{x}_2) = 1$. When the previous layer collapses the input points, the layer $l$ receives the same input for $\mathbf{x}_1$ and $\mathbf{x}_2$, resulting a perfect distance preservation as the distance between the same point is trivially zero. 

The following relationship between distortion and network distances is important.
\begin{equation}
\label{eq:distortion-distance-relationship}
\begin{split}
    d_{H_l}(h_{w_l}(\mathbf{x}_1), h_{w_l}(\mathbf{x}_2)) = \rho_l(\mathbf{x}_1, \mathbf{x}_2) * d_{H_{l-1}}(h_{w_{l-1}}(\mathbf{x}_1), h_{w_{l-1}}(\mathbf{x}_2))
\end{split}
\end{equation}

Equation~\ref{eq:distortion-distance-relationship} follows directly from the the fact that the input of a given layer $l$ is the output of the previous layer $h_w = h_{w_0}\circ h_{w_1}...\circ h_{w_L}$. When the previous layer does not collapse the input Equation~\ref{eq:distortion-distance-relationship} directly follows from the definition of $\rho_l$. In the case of feature collapse within the previous layer ($d_{H_{l-1}}(h_{w_{l-1}}(\mathbf{x}_1), h_{w_{l-1}}(\mathbf{x}_2)) = 0$), the input to the next layer is the collapsed point and $d_{H_l}(h_{w_l}(\mathbf{x}_1), h_{w_l}(\mathbf{x}_2)) = 0$ satisfying Equation~\ref{eq:distortion-distance-relationship}.

Using the network distortion coefficient we can rewrite the linear combination of distances as a function of the input distances $d_X$:

\begin{equation}
    \label{eq:intermediate-distance-derived}
    \begin{split}
        d_{SDN}(\Delta h(\mathbf{x}_1), \Delta h(\mathbf{x}_2)) & = \sum_{l=0}^{L}r_l d_{H_l}(h_{w_l}(\mathbf{x}_1), h_{w_l}(\mathbf{x}_2)),\\
        & = \sum_{l=0}^{L}r_l*d_X(\mathbf{x}_1, \mathbf{x}_2)* \prod_{i=0}^l \rho_i(\mathbf{x}_1, \mathbf{x}_2), \\
        & = d_X(\mathbf{x}_1, \mathbf{x}_2)*\sum_{l=0}^{L}r_l* \prod_{i=0}^l \rho_i(\mathbf{x}_1, \mathbf{x}_2), \\
        & = d_X(\mathbf{x}_1, \mathbf{x}_2)*C. \\
    \end{split}
\end{equation}

The first derivation follows from a recursive application of Equation~\ref{eq:distortion-distance-relationship}. The second, from the independence of $d_X(\mathbf{x}_1, \mathbf{x}_2)$ from both $i$ and $l$.

We note that Equation~\ref{eq:intermediate-distance-derived} satisfies Equation~\ref{eq:transitional-feature-preservation} when an appropriate weight choice $r_l$ results in $C=1$. A solution for $r_l$ only exists when the first layer is collapse resistant, i.e. when $\rho_0(\mathbf{x}_1, \mathbf{x}_2) \neq 0$ for $d_X(\mathbf{x}_1, \mathbf{x}_2) \neq 0$; a requirement for Proposition~\ref{prop:tfp-intermediate}.
\end{proof}

\subsection{Mutual Information of Intermediate Representations}
\label{app:mi-intermediate-representations}
In this section, we discuss how intermediate representations aid in increasing the information of the full input distribution within the uncertainty source representation. As discussed in Section~\ref{sec:dnn-information-plane}, effective uncertainty estimation is contingent on modelling information of the full input space $\mathcal{X}$ (not just $\mathcal{X}_{ID}$) to differentiate the training distribution from the test distribution. Within the context of the mutual information in neural networks \citep{tishby2000information}, achieving this is equivalent to maintaining the mutual information between the uncertainty source representation $Z$ (i.e. the representation used to compute the uncertainty $u(Z)$), and the input $X$. 

\begin{equation}
\label{eq:mutual-info-input-max}
\begin{split}
   I(Z; X).
\end{split}
\end{equation} 

A differentiator for uncertainty estimators is therefore their uncertainty source $Z$. We show that combining intermediate layers has favorable uncertainty properties in comparison to a conventional neural network output  - i.e. our method maintains the mutual information $I(Z; X)$ more effectively. 

In our algorithm, we measure uncertainty from a combination of intermediate layers $h^l_{w_l}$ instead of the final output. Within the context of mutual information, the joint representation $Z = h^1_{w_1}, ..., h^L_{w_L}$ maintains the following relationship for a layered neural network:

\begin{equation}
	\label{eq:mutual-information-intermediate}
	\begin{split}
		I( h^1_{w_1}, ..., h^L_{w_L}; X) \geq I(h_{w}(X); X) \\
	\end{split}
\end{equation}

The interpretation of Equation~\ref{eq:mutual-information-intermediate} is simple. Our method preserves information by extracting features before they are collapsed by subsequent network components. Hence, the mutual information with respect to the input is larger when intermediate representations are utilized in comparison to the final output exclusively. We further provide proof for Equation~\ref{eq:mutual-information-intermediate}:

\begin{proof}
	\label{proof:mutual-information-intermediate}
	For our discussion, we utilize data processing inequality \citep{einfo} within the context of neural networks. Specifically, given an intermediate layer $h^l_{w_l}$ the following relationship holds for any subsequent layers

	\begin{equation}
		\label{eq:data-processing-inequality}
		\begin{split}
			I(h^l_{w_l}(X); X) \geq I(h^{l+1}_{w_{l+1}}(X); X).
		\end{split}
	\end{equation}
	
	The mutual information of the joint variable $Z = h^1_{w_1}, ..., h^L_{w_L}$ and input $X$ can be expressed with the chain rule of mutual information

	\begin{equation}
		\label{eq:mutual-information-intermediate-derived}
		\begin{split}
			I(Z; X) & = I(h^1_{w_1}(X); X) - I(h^2_{w_2}(X), ..., h^L_{w_L}(X); X | h^1_{w_1}(X)) \\
			& = I(h^1_{w_1}(X); X) - H[X] + H[X | h^2_{w_2}(X), ..., h^L_{w_L}(X)] \\
			& = I(h^1_{w_1}(X); X) \\
			& \geq I(h_{w}(X); X) \\
		\end{split}
	\end{equation}
	
	The first derivation comes from the chain rule of mutual information, the seconde from the definition of mutual information, and the third from the fact that layered neural networks form a Markov chain with $X \rightarrow h^1_{w_1}(X) \rightarrow ... \rightarrow h^L_{w_L}(X)$ \citep{tishby2000information}. The final inequality is a direct manifestation of Equation~\ref{eq:data-processing-inequality}.
	
\end{proof}
\section{Impelementation Details}
In this appendix, we provide details of the different experimental setups and comparison methods used in this paper. All experiments are implemented with pytorch. When a implementation was publicly available, we heavily relied on it in our own code. This is the case for DUQ (\hyperlink{https://github.com/y0ast/deterministic-uncertainty-quantification}{https://github.com/y0ast/deterministic-uncertainty-quantification}), and SNGP (\hyperlink{https://github.com/google/uncertainty-baselines/blob/master/baselines/imagenet/sngp.py}{https://github.com/google/uncertainty-baselines/blob/master/baselines/imagenet/sngp.py}, as well as \hyperlink{https://github.com/y0ast/DUE}{https://github.com/y0ast/DUE}).

\subsection{Surface Plots and Class Distribution Experiments}
\label{app:exp-details-analysis}

\paragraph{Hyperparameter and Architecture Details}
In all experiments, we train a resnet-18 architecture \citep{he2016deep} over 200 epochs and optimize with stochastic gradient descent with a learning rate of 0.01. We further decrease the learning rate by a factor of 0.2 in epochs 100, 125, 150, and 175 respectively, and use the data augmentations random crop, random horizontal flip, and cutout to increase the generalization performance. For our experiments, we deploy direct spectral normalization of the convolutional, and batch normalization layers to implement representational feature preservation. On the full CIFAR100 dataset, we achieve an overall classification accuracy of 77.41 \% and 75.93 \% for the model with and without spectral normalization respectively. We average our results over three random seeds.

\begin{figure*}[!ht]
    \begin{center}
        \includegraphics[scale=0.5]{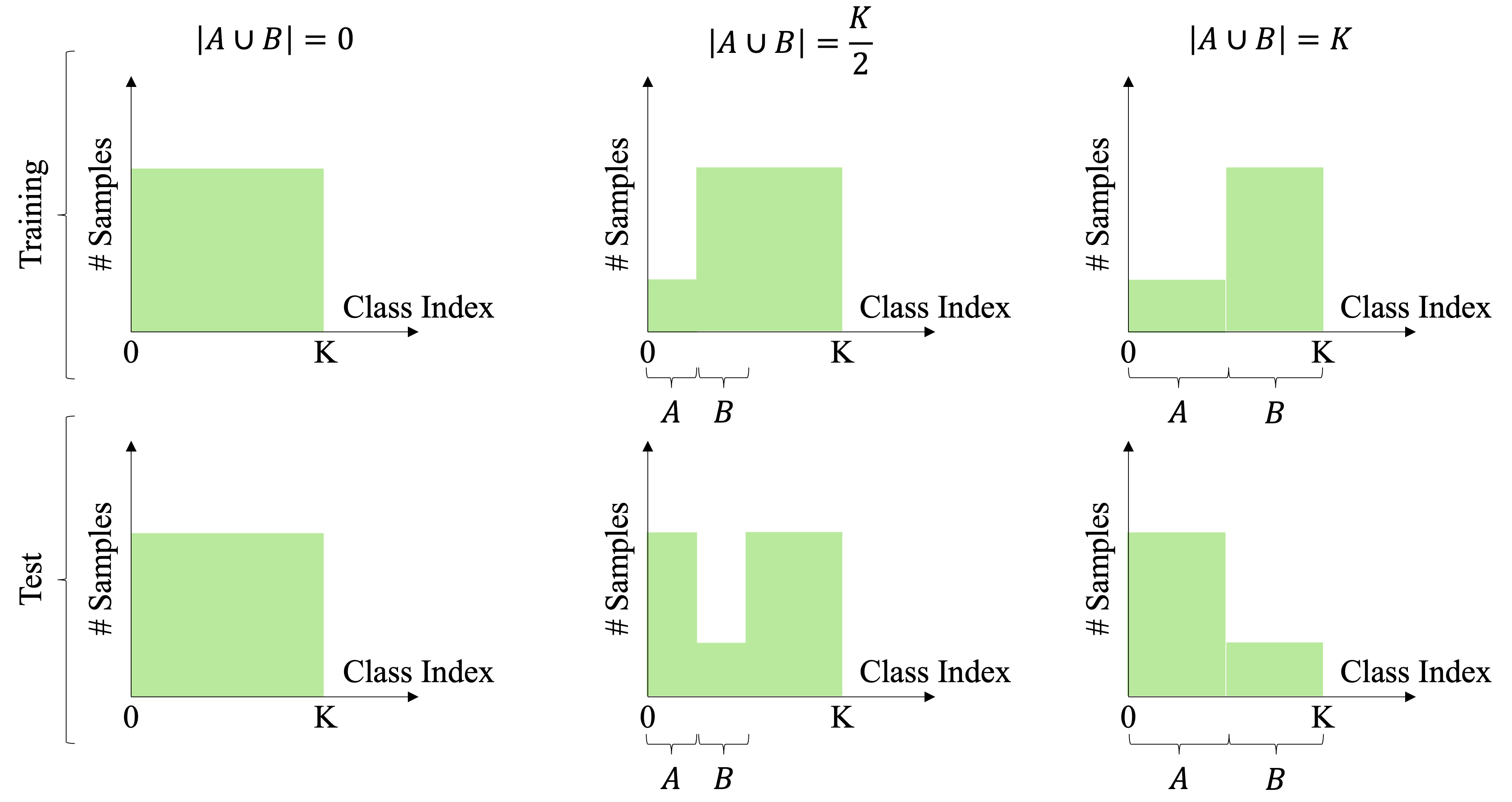}
    \end{center}
    \caption{Toy example of class distributions at different imbalance severity levels. Each column represents a different severity level, and each column the training and test set distribution respectively.}
    \label{fig:class-dist}
\end{figure*}

\paragraph{Imbalanced CIFAR100}
We imbalance the dataset as follows: for a certain subset of classes $A \subset \{1, ..., K\}$, we reduce the the number of training samples by 80\% and do not change the number of test samples. For a second subset $B \subset \{1, ..., K\}, A \cap B = \emptyset$, we reduce the number of test samples by 90\% and do not change the number of training samples. As a result, the first subset of classes contains few training samples and a large amount of test samples, while the other set suffers from the opposite problem. The imbalance severity can be adjusted by the number of classes in both imbalanced subsets $A \cup B \subset \{1, ..., K\}$. For simplicity, we keep the same amount of classes in both subsets $A$ and $B$. In Figure~\ref{fig:class-dist}, we show an overview of class distributions at different severity levels. Here, the top and bottom row contain the training and test distribution respectively. We perform our entire analysis on the CIFAR100 dataset as it represents a challenging benchmark with a large class variety. In addition, the dataset is fully balanced and contains the same class distribution for training and testing.

\subsection{Ablation Study}
We further provide an ablation study on utilizing different constellations of \texttt{TULIP}. We investigate equal weighting of the individual intermediate scores, as well as removing the GP layer. We show our results in Table~\ref{tab:ablation}. As shown, the final version of \texttt{TULIP} outperforms the other constellations further supporting the algorithm.

\begin{table}[h]
\scriptsize
\centering
\caption{Weight Ablation Study.}
\label{tab:ablation}

\begin{tabular}{l c c c}
\hline
              Algorithms & CIFAR10-C & CIFAR100-C & SVHN \\
		\hline
		\hline
        \texttt{TULIP} - Equal Weighting   & 0.716 ± 0.006 & 0.883 ± 0.009 & 0.840 ± 0.023    \\
    	\texttt{TULIP} - No GP, Switch Weighting  & 0.722 ± 0.005 & 0.885 ± 0.008 & 0.811 ± 0.031      \\
		\texttt{TULIP}     & 0.738 ± 0.006   & 0.936 ± 0.003 & 0.946 ± 0.012   \\
\end{tabular}

\end{table}

\subsection{Comparison Method Details}
\label{app:com-method-details}
In the following, we provide details on each feature preservation method.

\begin{itemize}
    \item \emph{Energy-Based Model} \citep{liu2021energybased}: for the energy based model, uncertainty estimates are derived by replacing the final softmax layer with the unnormalized softmax density. No additional feature preservation constraint is used during training. In our experiments, we only compare against the version that does not require additional out-of-distribution data.
    \item \emph{DUQ} \citep{van2020duq}: in DUQ, uncertainty is inferred by the closest kernel distance. To ensure the preservation of features, DUQ implements the double sided gradient penalty, penalizing the squared distance of the gradient from a fixed value at every input point. In contrast to spectral normalization, it implements feature preservation with a \emph{local} constraint, without explicit guarantees for data points outside of $\mathcal{X}_{ID}$.
    \item \emph{SNGP} \citep{liu2020simple}: SNGP infers uncertainty through distance awareness within the output. In the original paper, this is achieved by replacing the final output layer with a gaussian process approximation, and implementing spectral normalization in combination with residual layers. Direct spectral normalization on the weights provides a upper lipschitz bound while the combination with residual connections further ensures a lower lipschitz bound on the distance between two input points. Both bounds jointly result in feature preservation, as distances in between input points are approximately preserved when traversing the network. In comparison to the gradient penalty, spectral normalization enforces a \emph{global} constraint. 
\end{itemize}

\subsection{Out-of-Distribution Experiments on CIFAR10/100}
\label{app:ood-c10/100}
In our out-of-distribution experiments, we use the same backbone residual architectures (ResNet-50, -101, and -152) with a batch size of 128. For all setups, we use the standard data augmentations random horizontal flip, random crop, and cutout. In the following, we describe the details for each method. All Results are averaged over three random seeds.

\paragraph{Softmax DNN and Energy-Based Model}
We train both models with the SGD optimizer and an initial learning rate of 0.01. We optimize the model for 200 epochs and reduce the learning rate by a factor of 0.2 in epochs 100 and 150. For the energy-based model, we use the unnormalized softmax density, similar to other implementations \citep{mukhoti2023deep}.

\paragraph{DUQ}
Our DUQ models are trained with the SGD optimizer and a learning rate of 0.05. We train for 600 epochs and reduce the learning rate by factor 0.2 in epochs [300, 375, 450, 525]. For the gradient penalty weight, we perform the experiment with hyperparameters from the original paper \citep{van2020duq}, as well as a newer implementation from \cite{postels22apracticality}, and report the constellation with the highest accuracy value.

\paragraph{SNGP}
We trained SNGP with the SGD optimizer and an initial learning rate of 0.01. We reduce the learning rate by a factor of 0.2 in epochs 100 and 150, and train for 200 epochs. We further use a spectral normalization coefficient of three.

\paragraph{Our Method}
We train each SDN model with the SGD optimizer using an initial learning rate of 0.01. Further, we optimize the architecture for 400 epochs and reduce the learning rate by a factor of 0.2 in epoch 200, and 300. The architecture of the internal classifiers is similar to \cite{kaya2019shallow}, with a single linear layer combined with a mixture of average-/max-pooling where necessary. The output layer is then fed into a GP layer, which has the same architecture as SNGP. We distribute the internal classifiers equally distanced across the network by placing a internal classifier on top of every third residual block. For ResNet-50 this is equivalent to every sixth layer, and every ninth layer for the remaining larger models. Our selection is geared towards simplicity and performance may be further improved with other uncertainty scores such as entropy or energy functions. We train each model with a equally weighted SDN loss. We set the switch parameter $N_s$ to the average number of switches occurring in the validation set as a general rule, but set it to 1 for the small scale datasets where few switches occur and resulting in a negligible switch mean. 


\subsection{ImageNet}
\label{app:imgnet}
For our experiments on ImageNet, we train the baseline DNN with standard training settings similar to baseline recipe in \href{https://pytorch.org/blog/how-to-train-state-of-the-art-models-using-torchvision-latest-primitives/}{https://pytorch.org/blog/how-to-train-state-of-the-art-models-using-torchvision-latest-primitives/}. For our MC-Dropout implementation, we place one dropout layer in between every residual bottleneck block. We set the dropout probability to 0.01 when deriving uncertainty estimates. For our SDN implementation, we use the same number of internal classifiers as in our previous experiments and set $N_s$ to the mean number of switches in the validation set. We choose this setting due to simplicity and it allows us to choose hyperparameters in an automated fashion. We train our model with an equally weighted SDN loss and train for 100 expochs with an adam optimizer and a learning rate of 0.0001.

\subsection{Artificial Dataset}
\label{app:exp-setup-art-dataset}
For our experiments in Section~\ref{app:disagreement-weight-tuning}, we train our models on an artificial spiral dataset with three different classes. Here, each spiral arm represents a class that starts in the center, and spirals for one full loop of 360°. In our ensemble experiments, we use a three-layer MLP architecture in an ensemble of ten. During training, we use a SGD optimizer with a learning rate of 0.008 and train each ensemble element for 400 epochs. To measure disagreement in between layers, we adjust the MLP architecture into a SDN, by placing an internal classifier on the first, and second layer respectively. We train the model with an SDN loss as described in Equation~\ref{eq:sdn-loss}, optimize with the adam variant of SGD, and select a learning rate of
0.001. Due to the higher loss complexity, we train the SDN model for 800 epochs. To further measure disagreement, we utilize the same measure as previous works \cite{mukhoti2023deep, malinin2019ensemble}.

\begin{equation}
	\label{eq:disagreement-mi}
	\begin{split}
		disagreement(\mathbf{x}) = H[\frac{1}{N_{IC/E}} \sum_{i=1}^{N_{IC/E}} p(y | \mathbf{x}, w_i)] - \frac{1}{N_{IC/E}} \sum_{i=1}^{N_{IC/E}} H[p(y | \mathbf{x}, w_i)].
	\end{split}
\end{equation}

Here, $N_{IC/E}$ denotes the number of internal classifiers or ensembles respectively, and $p(y | \mathbf{x}, w_i)$ the target posterior distribution of the individual model elements.

\section{Additional Experiments}
\subsection{Surface Plots for Distance Preservation under Class Imbalance}
\label{app:uncertainty-surface}
\begin{figure*}[!ht]
    \begin{center}
        \includegraphics[scale=0.5]{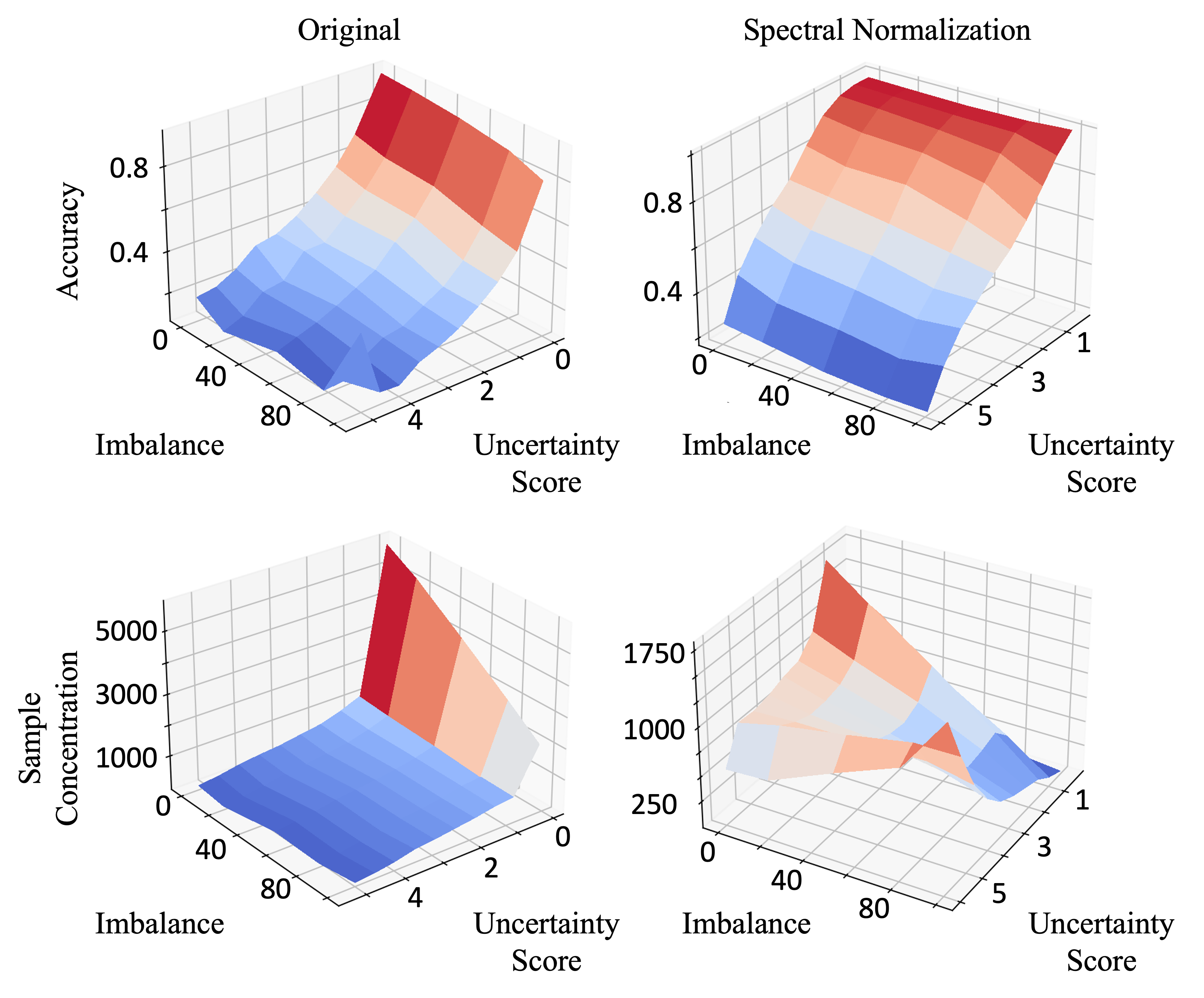}
    \end{center}
    \caption{Accuracy and number of samples (sample concentration) with respect to dataset imbalance and average uncertainty score. In the the top row we show accuracy, in the bottom row we show sample concentration. The left column represents a conventional DNN while the right shows feature distance preservation through spectral normalization.}
    \label{fig:surface-plots}
\end{figure*}

In addition to class accuracy, we further wish to analyze uncertainty estimates under class imbalance. For this purpose, we plot both accuracy and the number of samples (sample concentration) with respect to imbalance severity and uncertainty scores (Figure~\ref{fig:surface-plots}). The accuracy plots provide information of the calibration capabilities in relation to class imbalance. Ideally, the uncertainty fully informs of the accuracy of a sample and the dependency is linear on the y-/z-plane \citep{guo2017calibration}. We note that a conventional neural network is not calibrated and overconfident in its prediction - the dependency is not linear. Spectral normalization significantly improves along this characteristic (top right plot), and improves linearity regardless of the imbalance severity. The bottom row complements our accuracy curves. For low imbalance severities the majority of samples concentrate low uncertainty/high accuracy regions on the x-/y-plane. However, the dependency inverts with increasing imbalance. Samples concentrate in high-uncertainty/low-accuracy regions complementing the accuracy decline in Figure~\ref{fig:acc-analysis}.
\subsection{Disagreement Analysis of Proxy Labels}
\label{app:disagreement-weight-tuning}
\begin{figure*}
    \begin{center}
        \includegraphics[scale=0.27]{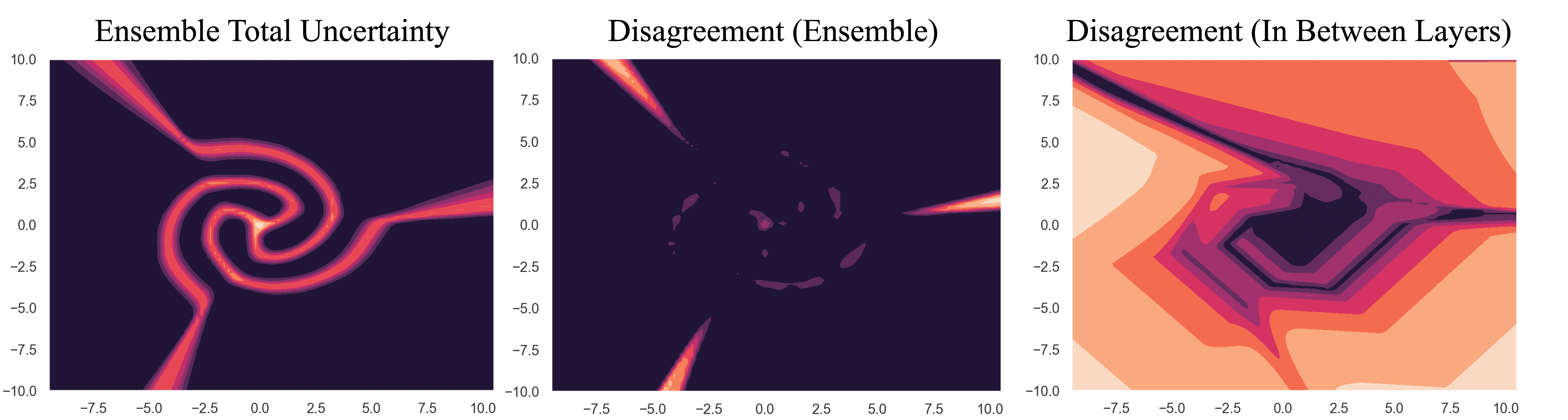}
    \end{center}
    \caption{Comparison of disagreement in ensembles and intermediate representations trained on an artificial spiral dataset. The left image represents the total uncertainty of an ensemble of ten. In the center, we display model disagreement within the ensembles. The right depicts the disagreement within the intermediate representations of a single deterministic neural network. Dark red, or black represents low uncertainty while lighter shades of red or orange depict the opposite.}
    \label{fig:disagreement-ensemble-intermediate}
\end{figure*}

In this section, we provide a detailed analysis on the uncertainty proxy scores used in \texttt{TULIP}. In our analysis, we compare the properties of intermediate representation with ensembles uncertainty scores and analyze the disagreement among both methods. Within the context of ensembles, disagreement is used to derive the difference or ``spread" of ensembles and is frequently used directly as an uncertainty score in several contexts \citep{malinin2019ensemble}. To showcase disagreement within intermediate representations, we compare against ensembles on an artificial spiral dataset (Figure~\ref{fig:disagreement-ensemble-intermediate}). The left image shows the total uncertainty of an ensemble of ten and is among the most common usages of ensembles. The center shows disagreement among the different ensemble models as derived by previous work \citep{malinin2019ensemble}, and approximates uncertainty occurring due to data scarcity \cite{kendall2017uncertainties}. Ideally, the measure is low where sufficient data samples are available (center of the spiral), and increases where little or no data is available. The right shows the same measure of disagreement, with the exception of measuring in between individual SDN outputs instead of ensemble models. We note, that ensembles exhibit high disagreement near the decision boundaries exclusively, while the SDN model comprehensively approximates data scarcity in between layers. Our observations can be explained along the intuition of feature preservation. Ensembles measure disagreement among the output of the entire network architectures, and collapse important information from the input distribution that can be leveraged for uncertainty estimation. In contrast, intermediate representations contain more information of the input distribution and provide a coarse measure of data scarcity. Overall, we gather that disagreement in between internal classifiers is a reasonable approach to determine the hyperparameters $r_i$. Experimental details to produce Figure~\ref{fig:disagreement-ensemble-intermediate} can be found in Appendix~\ref{app:exp-setup-art-dataset}.

\subsection{Additional Architectures}
We further compare \texttt{TULIP} with additional architectures. Within this context, we explore both the vision transformer \cite{visiontransformer}, as well as the convmixer \cite{convmixer}. For the vision transformer, we use a similar setup to our imagenet experiments in Section~\ref{app:imgnet-results}. For the convmixer, we investigate medical modalities similar to Section~\ref{sec:medical-modalities}. We show our results in Table~\ref{tab:imgnet-vit} and Table~\ref{tab:organ-ood-convmixer}. Similar to our previous experiments, \texttt{TULIP} overwhelmingly matches or outperforms the comparison methods supporting our approach with transitional feature preservation.

\begin{table*}
\tiny
\centering
\caption{OOD detection and classification accuracy on the ImageNet dataset.}
\label{tab:imgnet-vit}
\begin{tabularx}{\textwidth}{X c c c}
\hline
              Algorithms & Accuracy & ImageNet-C \\
		\hline
		\hline
        DNN                & \multirow{3}{*}{78.546 ± 0.000}       & 0.822 ± 0.000    \\
    	Energy-Based\citep{liu2021energybased}       &     & 0.819 ± 0.000      \\
    	MC-Dropout\citep{gal2016dropout}         &     & 0.839 ± 0.001     \\
        \hline
		\texttt{TULIP}    & 78.546 ± 0.000   & \bf{0.851 ± 0.004}   \\
\end{tabularx}

\end{table*}
\begin{table*}
\centering
\caption{AUROC and classification accuracy on on the organ\{C, A, S\} datasets.}
\label{tab:organ-ood-convmixer}
\tiny
\begin{tabularx}{\textwidth}{l X c c c c c}
\hline
              &           &  &        & \multicolumn{3}{c}{\centering AUROC} \\
              Dataset & Algorithms & Preservation Constraint & Accuracy & organA & organC & organS \\
		\hline
		\hline
        \multirow{4}{*}{organA}
        & DNN               & \multirow{2}{*}{No Constraint} & \multirow{2}{*}{93.717 ± 0.197}       & -      & \bf{0.904 ± 0.003}      & \bf{0.904 ± 0.003}        \\
    	& Energy-Based \citep{liu2021energybased}      & &     & -      & \bf{0.894 ± 0.004}      & 0.847 ± 0.002       \\
    	& SNGP\citep{liu2020simple} & SN & 89.810 ± 0.000       & -      & 0.834 ± 0.015      & 0.826 ± 0.014      \\
        \cline{2-7}
    	& \texttt{TULIP}    &  No Constraint  & \bf{94.880 ± 0.324}       & -      & \bf{0.897 ± 0.013}      & \bf{0.855 ± 0.006}        \\
		\hline
		\hline
        \multirow{4}{*}{organC}
        & DNN               & \multirow{2}{*}{No Constraint} & \multirow{2}{*}{90.221 ± 0.018}       & \bf{0.836 ± 0.007}      & -      & 0.762 ± 0.006        \\
    	& Energy-Based \citep{liu2021energybased}  &   &     & 0.803 ± 0.015      & -      & 0.738 ± 0.010       \\
    	& SNGP\citep{liu2020simple} & SN & 87.530 ± 0.000       & 0.755 ± 0.000      & -      & 0.729 ± 0.000      \\
        \cline{2-7}
    	& \texttt{TULIP}     &  No Constraint  & \bf{90.941 ± 0.109}       & \bf{0.842 ± 0.001}      & -      & \bf{0.772 ± 0.002}        \\
		\hline
		\hline
        \multirow{4}{*}{organS}
        & DNN               & \multirow{2}{*}{No Constraint} & \bf{\multirow{2}{*}{80.672 ± 0.085}}       & 0.688 ± 0.028      & 0.733 ± 0.010      & -        \\
    	& Energy-Based \citep{liu2021energybased}     & &     & 0.720 ± 0.028      & 0.777 ± 0.014      & -       \\
    	& SNGP\citep{liu2020simple} & SN & 70.671 ± 0.980       & 0.640 ± 0.019      & 0.713 ± 0.007      & -      \\
        \cline{2-7}
    	& \texttt{TULIP}      &  No Constraint  & 78.984 ± 0.017       & \bf{0.757 ± 0.037}      & \bf{0.794 ± 0.017}      & -        \\
\end{tabularx}

\end{table*}
\subsection{Calibration and Runtime}
\label{app:calibration-experiments}
We further investigate calibration as characteristic for uncertainty score quality. With calibration, we refer to the capability of the output score to be reflective of the actual generalization performance \citep{guo2017calibration}. For this purpose, we consider the expected calibration error (ECE) \citep{naeini2015obtaining} to measure miscalibration and show our results in Figure~\ref{tab:ece-rn}. Our setup is equivalent to our experiments on out-of-distribution detection and we further show the runtime normalized by the latency of a conventional DNN. Specifically, we measure the latency of a single batch for each model and divide by the latency of a conventional DNN. For all of our experiments we use a single NVIDIA GeForce GTX 1080 Ti. We note, that all algorithms are equivalent in terms of runtime and that our method matches or outperforms comparable single-pass methods in the majority of benchmarked constellations. Our results show both the runtime benefits and a high uncertainty estimation quality for our method.

\begin{table*}
\tiny
\centering
\caption{Expected-Calibration-Error and Runtime on CIFAR10 and CIFAR100.}
\label{tab:ece-rn}
\resizebox{\textwidth}{!} {
\begin{tabular}{l l c c c c}
\hline
                         &          & & & \multicolumn{2}{c}{\centering ECE} \\
              Architecture & Algorithms & Runtime & Space Complexity & CIFAR10 & CIFAR100 \\
		\hline
		\hline
        \multirow{7}{*}{ResNet-50}
        & DNN               & \multirow{2}{*}{1x}  & \multirow{2}{*}{1x}     & \multirow{2}{*}{0.065 ± 0.001}    & \multirow{2}{*}{1.110 ± 0.002}   \\
    	& Energy-Based \citep{liu2021energybased}      &    &    &    \\
    	& SNGP\citep{liu2020simple} &   1x   &   1x   & \bf{0.010 ± 0.002}      & 0.720 ± 0.002        \\
		& DUQ\citep{van2020duq}     &   1x   &   1x   & 0.973 ± 0.003      & -       \\
        \cline{2-6}
    	& \texttt{TULIP}         &   1x   &   1.5x   & 0.030 ± 0.005      & \bf{0.658 ± 0.004}             \\
		\hline
		\hline
        \multirow{5}{*}{ResNet-101}
        & DNN                &  \multirow{2}{*}{1x}   &  \multirow{2}{*}{1x}   & \multirow{2}{*}{0.056 ± 0.001}      & \multirow{2}{*}{1.108 ± 0.011}            \\
    	& Energy-Based \citep{liu2021energybased}      &   &   &          \\
    	& SNGP\citep{liu2020simple} &   1x   &   1x   & 0.039 ± 0.013      & \bf{0.671 ± 0.021}     \\
		& DUQ\citep{van2020duq}     &   1x   &   1x   & 0.973 ± 0.001      & -    \\
        \cline{2-6}
    	& \texttt{TULIP}         &   1x   &   1.5x   & \bf{0.025 ± 0.021}    & \bf{0.672 ± 0.031}        \\
     \hline
	 \hline
        \multirow{5}{*}{ResNet-152}
        & DNN               &   \multirow{2}{*}{1x}  &   \multirow{2}{*}{1x}   & \multirow{2}{*}{0.055 ± 0.001}      & \multirow{2}{*}{1.126 ± 0.004}     \\
    	& Energy-Based \citep{liu2021energybased}      &    &   & \\
    	& SNGP\citep{liu2020simple} &   1x   &   1x   & 0.050 ± 0.008      & \bf{0.698 ± 0.005}   \\
		& DUQ\citep{van2020duq}     &   1x   &   1x   & 0.623 ± 0.002      & -   \\
        \cline{2-6}
    	& \texttt{TULIP}         &   1x   &   1.5x   & \bf{0.026 ± 0.011}    & \bf{0.662 ± 0.055}       \\
\end{tabular}
}

\end{table*}
\subsection{Imbalanced Out-of-Distribution Experiments}
\label{app:imbalanced-ood}
In addition to standard out-of-distribution detection, we consider less informative training sets in the form of class imbalance. For this purpose, we unbalance the full 100 classes of the CIFAR100 dataset, as described in our previous analysis in Section~\ref{sec:theory-overall}. For the first 50 classes we reduce the training samples by 80\% (400 samples) and maintain the same test set. For the remaining 50 classes, we reduce the test set by 90\% (90 samples) and maintain all training samples. We use the same implementations as our previous out-of-distribution experiments and consider the three residual backbones ResNet-50, -101, and -152. We show the AUROC scores in Table~\ref{tab:ood-c100-imbalanced}. Complementary to our previous results, our method outperforms other single-pass uncertainty estimators despite having access to only 90\% of the training labels. These results illustrate the importance of using feature preservation methods that do not oppose the training objective and show that intermediate representations are attractive options for maintaining information of the input distribution.

\begin{table*}
\tiny
\centering
\caption{OOD detection and classification accuracy on the imbalanced CIFAR100 dataset vs. SVHN.}
\label{tab:ood-c100-imbalanced}
\resizebox{\textwidth}{!} {
\begin{tabular}{l l c c}
\hline
              Architecture & Algorithms & Accuracy & AUROC \\
		\hline
        \multirow{5}{*}{ResNet-50}
        & DNN                                      & 50.630 ± 0.155  & 0.747 ± 0.019       \\
        & Energy-Based \citep{liu2021energybased}  & 50.630 ± 0.155       & 0.772 ± 0.042       \\
    	& SNGP\citep{liu2020simple}                & 45.752 ± 3.375 & 0.783 ± 0.041     \\
		& DUQ\citep{van2020duq}                    & -       & -     \\
        \cline{2-4}
    	& \texttt{TULIP}                & \bf{52.388 ± 0.568}  & \bf{0.840 ± 0.019}        \\
     \hline
     \hline
        \multirow{5}{*}{ResNet-101}
        & DNN                                      & 49.782 ± 0.161  & 0.751 ± 0.006       \\
        & Energy-Based                             & 49.782 ± 0.161  & 0.793 ± 0.003 \\
        & SNGP\citep{liu2020simple}                & 44.073 ± 2.186       & 0.748 ± 0.018     \\
        & DUQ\citep{van2020duq}                    & -       & -     \\
        \cline{2-4}
    	& \texttt{TULIP}                & \bf{52.648 ± 1.027}  & \bf{0.807 ± 0.028}          \\
\end{tabular}
}
\end{table*}

\section{Method Details}
\label{app:summary-method}
In our implementation, we calculate the individual uncertainty scores through distance awareness, similar to SNGP \citep{liu2020simple}. Applied top our algorithm, the final layer of both the internal classifier, as well as the prediction output are Laplace-approximated Gaussian processes, and we calculate uncertainty with the Dempster-Shafer metric:

\begin{equation}
	\label{eq:sdn-subgroup-logregression}
	\begin{split}
		u_s(\mathbf{x}) = \frac{K}{K + \sum_{k=1}^K exp(g^k(\mathbf{x}))}
	\end{split}
\end{equation}

Here, $g^k$ is the k-th logit of the output $g$ (either model prediction or internal classifier), and $K$ represents the number of classes. Our choice is based on simplicity and the past success of Gaussian process layers in single-pass uncertainty estimation \citep{van2021due, liu2020simple}. While our design produces sufficient results, we emphasize that other implementations of both $u_s$ and $h_w$ may further improve the performance.

\begin{minipage}{0.5\textwidth}
\begin{algorithm}[H]
    \centering
    \caption{Training}\label{alg:turing-training}
    \begin{algorithmic}[1]
        \STATE $\mathbf{Input}$ 
        \STATE \text{Labeled Training Set }
        \STATE $\{ \mathbf{x}_i \in \mathcal{X}_{ID}, y_i\}_{i = 1}^N$
        \STATE \text{Unlabeled Validation Set }
        \STATE $\{ \mathbf{x}_i \in \mathcal{X}_{ID}\}_{i = 1}^{N_{val}} $ 
        \item[] 
        \STATE \textbf{SDN Training}
        \FOR{$epoch \in [1, epochs]$} 
        \STATE $h_w \gets \text{SGD Update } L_{SDN}$
        \ENDFOR
        \item[] 
        \STATE \textbf{Fit Combination Head}
        \STATE Derive prediction switches and scores
        \STATE $\mathbf{s} \gets \{ s(\mathbf{x}_{i}) \}_{i=1}^{N_{val}}$ 
        \STATE $\mathbf{v} \gets \{ [u_s(f^1_{w_1}(\mathbf{x}_i)), .., u_s(f^{N_{IC}}_{w_{N_{IC}}}(\mathbf{x}_i))] \}_{i=1}^{N_{val}}$ 
        \STATE Fit logistic regression weights
        \STATE $r_1, ..., r_{N_{IC}} = LR(\mathbf{s}, \mathbf{v})$
        \STATE \textbf{Return}  $h_w, r_1, ..., r_{N_{IC}}$
    \end{algorithmic}
\end{algorithm}
\end{minipage}
\hfill
\begin{minipage}{0.46\textwidth}
\begin{algorithm}[H]
    \centering
    \caption{Prediction}\label{alg:turing-inference}
    \begin{algorithmic}[1]
        \STATE $\mathbf{Input}$ 
        \STATE $\text{Test Sample } \mathbf{x}_{te}$
        \item[] 
        \STATE \textbf{SDN Prediction}
        \STATE Internal uncertainty scores
        \STATE $\mathbf{u} \gets [u_s(f^1_{w_1}(\mathbf{x}_{te})), ..., u_s(f^{N_{IC}}_{w_{N_{IC}}}(\mathbf{x}_{te}))]$
        \item[] 
        \STATE Combine scores
        \STATE $u_{final} \gets \frac{1}{\sum_{i=1}^{N_{IC}} r_i} \sum_{i=1}^{N_{IC}} r_i u_s(f^i_{w_i}(\mathbf{x}))$
        \item[] 
        \STATE Prediction
        \STATE $\Tilde{y}_{te} \gets h_w(\mathbf{x}_{te})$
        \STATE \textbf{Return}  $\Tilde{y}_{te}, u_{final}$
    \end{algorithmic}
\end{algorithm}
\end{minipage}

\section{Additional Related Work}
\label{app:distance-preserving}
\paragraph{Distance Preserving Neural Networks}
The goal of learning a distance-preserving mapping has been an important objective in a wide range of fields such as generative modeling \citep{lawrence2006, dinh2014nice, dinh2016density} and dimensionality reduction \citep{abraham2006advances, perrault2012metric}. Recently, the concept has been expanded to uncertainty estimation for neural networks and is used to enable single-pass uncertainty estimators \citep{liu2020simple}. Several methods exist to control distance preservation in neural networks and each comes with its own set of trade-offs: the two-sided gradient penalty \citep{gulrajani2017improved} was originally introduced in the context of GANs as an alternative to weight clipping  \citep{arjovsky2017wasserstein}. The penalty regularizes the network by penalizing the squared distance of the gradient from a fixed value for every input point. The approach is popular due to its simple implementation, but represents only a soft constraint. Spectral normalization \citep{gouk2021regularisation, miyato2018spectral} combines spectral normalization with residual connections to implement distance preservation. The method represents a global constraint as it regularizes by normalizing the weights and suffers less from training instabilities as the gradient penalty. Finally, there exists work on reversible networks that force distance preservation through reversible layers and avoiding down-scaling operations \citep{jacobsen2018revnet, behrmann2019invertible}. In practice, reversible models are difficult to train and consume considerably more memory in practice \citep{van2021due}. For this purpose, recent single-pass approaches utilize either spectral normalization, or the gradient penalty.

\paragraph{Iterative Uncertainty Estimation}
In addition to single-pass uncertainty, significant related work exists in iterative uncertainty estimation. With iterative uncertainty estimation, we refer to methods requiring several forward passes for computation. Here, the state-of-the-art are deep ensembles \citep{lakshminarayanan2017simple}, as well as several parameter-efficient counterparts \citep{wen2020batchensemble, dusenberry2020efficient, thiagarajan2022single}. Further examples include Bayesian Neural Networks \citep{wenzel2020good, osawa2019practical} and MC-Dropout \citep{gal2016dropout}. In practice, these methods tend to render powerful uncertainty estimates but require several forward passes to compute, limiting their applicability in practice. In Appendix~\ref{app:distance-preserving}, we provide additional related work on distance preserving neural networks.



\end{document}